\documentclass[12pt]{article}

\usepackage[colorlinks,
            linkcolor=blue,
            anchorcolor=blue,
            citecolor=blue]{hyperref}
\usepackage{graphicx}
\usepackage{booktabs}
\usepackage{threeparttable}
\usepackage{subfigure}
\usepackage{footnote}
\usepackage{tabu}
\usepackage{tabularx}

\usepackage{amsmath}
\usepackage{amsthm}
\usepackage{algorithmic}

\usepackage[final]{changes} 
\definechangesauthor[name={Q.L.}, color=orange]{Q.}

\usepackage{subfigure}
\usepackage[ruled]{algorithm2e}
\usepackage{float}

\usepackage[numbers,sort&compress]{natbib}

\usepackage{subfigure}
\makeatletter
\renewcommand{\@thesubfigure}{\normalsize(\textbf{\alph{subfigure}})}
\makeatother

\usepackage{geometry}
\def\bq{\begin{equation}}
\def\eq{\end{equation}}
\def\bea{\begin{eqnarray}}
\def\eea{\end{eqnarray}}

\title{Improving Image Clustering through Sample Ranking and Its Application to remote--sensing images}
\author{Qinglin Li$^{1,2}$, Guoping Qiu$^{1,2,3,4,*}$\\
$^{1}$ College of Electronic and Information Engineering, Shenzhen University,\\ Shenzhen 518052, China; qlilx@szu.edu.cn (Q.L.);\\
$^{2}$  Guangdong Key Lab for Intelligent Information Processing, \\ Shenzhen University, Shenzhen 518052, China.\\
$^{3}$ Shenzhen Institute of AI and Robotics for Society, \\Shenzhen 518172, China\\
$^{4}$ Pengcheng Laboratory, Shenzhen 518055, China\\
$^{*}$Correspondence: qiu@szu.edu.cn
}

\date{}

\begin{document}
\maketitle

\begin{abstract}

Image clustering is a very useful technique that is widely applied to various areas, including remote sensing. Recently, visual representations by self-supervised learning have greatly improved the performance of image clustering. To further improve the well-trained clustering models, this paper proposes a novel method by first ranking samples within each cluster based on the confidence in their belonging to the current cluster and then using the ranking to formulate a weighted cross-entropy loss to train the model. For ranking the samples, we developed a method for computing the likelihood of samples belonging to the current clusters based on whether they are situated in densely populated neighborhoods, while for training the model, we give a strategy for weighting the ranked samples. We present extensive experimental results that demonstrate that the new technique can be used to improve the State--of--the--Art image clustering models, achieving accuracy performance gains ranging from $2.1\%$ to $15.9\%$. Performing our method on a variety of datasets from remote sensing, we show that our method can be effectively applied to remote--sensing images. The paper has been published on Remote Sensing \url{https://doi.org/10.3390/rs14143317}. 

\end{abstract}

\section{Introduction}


Clustering remote--sensing images containing objects belonging to the same class is an essential topic in the context of Earth observation. Without manual annotation, estimating the semantic similarities of images is difficult, and thus, clustering images is still a very challenging task. 
Images are usually described in very high-dimensional spaces. To effectively measure the similarities among images, it is necessary to find their low-dimensional representations. 

Deep supervised learning \cite{goodfellow2016deep,lecun2015deep,bengio2013representation} possesses a distinguished capability to extract good representations, and remarkable progress has been achieved in the past decade. 
In recent years, deep unsupervised or self-supervised representation learning has been developing rapidly, and various successful methods have been proposed, mainly including: pretext tasks \mbox{\cite{conf/iccv/DoerschGE15,conf/eccv/NorooziF16,conf/wacv/KimCYK18,conf/iclr/GidarisSK18,conf/cvpr/KolesnikovZB19,conf/eccv/ZhangIE16,conf/eccv/LarssonMS16,conf/iccv/WangG15}}, contrastive learning \cite{journals/corr/abs-1807-03748, conf/cvpr/YeZYC19,journals/corr/abs-1905-09272,conf/eccv/TianKI20,conf/cvpr/He0WXG20,journals/corr/abs-2002-05709}, generative models \cite{henaff2019dataefficient,HintonETAL:06,conf/icml/LeeGRN09,conf/cvpr/TangSH12,JMLR:v11:vincent10a,conf/cvpr/RenL18,conf/cvpr/JenniF18,conf/nips/XieDDHN17,conf/nips/DonahueS19}, clustering--based methods \mbox{\cite{conf/eccv/CaronBJD18,conf/icml/HuangDGZ19,conf/iclr/AsanoRV20a,li2021designing},} and so on.
Good representations can be applied to various downstream tasks, and one of them is unsupervised classification or clustering. Using representations from self-supervised learning, SCAN~\cite{conf/eccv/GansbekeVGPG20} obtained the neighbor relations of images, which greatly improved the performances of image clustering and unsupervised classification, which can be further enhanced by self--training \cite{rosenberg2005semi,lee2013pseudo,xie2020self,sohn2020fixmatch} (using the model's prediction as pseudo labels to train the model via standard cross-entropy loss).


However, as learning is performed without supervision, it is inevitable that samples from multiple classes will be grouped under the same cluster. Existing methods such as SCAN~\cite{conf/eccv/GansbekeVGPG20} address this problem by only retaining the highest confidence predictions of the model as pseudo labels for training and using strong augmentations to prevent the model from over-fitting. Such solutions \cite{conf/eccv/GansbekeVGPG20,park2021improving,niu2021spice} throw away not only incorrectly clustered samples, but also those correctly clustered. This would hurt the data diversity \added[comment={added}]{\cite{gong2019diversity}}, which cannot be obtained from augmentations. Furthermore, samples in class boundaries are very valuable in clustering or classification, as is well known in the active learning \added[comment={added}]{ \cite{settles2009active,settles2012active}} and support vector machines \added[comment={added}]{ \cite{hearst1998support}} literature. On the other hand, keeping inappropriate pseudo labels would mislead model training \added[comment={added}]{ \cite{zhou2018brief,frenay2013classification,angluin1988learning,gao2016risk}}. Instead of throwing away huge amounts of samples, we make use of most samples in training by developing a method to rank the samples in each cluster based on their confidence of belonging to the cluster: samples with a higher confidence are assigned a larger weight, whilst those with a lower confidence are assigned a smaller weight in the cross-entropy loss.

An image cluster formed by various clustering algorithms is likely to have a dominant class, i.e., one class will have much more samples than other classes within each cluster. Within a cluster, samples from the dominant class can be regarded as the signal and samples from other classes as noise. It is reasonable to assume that signal samples are more likely to belong to the current cluster and, therefore, should be kept, while noise samples should be re-assigned to other clusters. We propose a statistical method to estimate the likelihood that a sample is signal or noise based on whether it is situated in a densely populated region. By ranking the samples within a cluster based on this likelihood, we propose a method to improve image clustering by manipulating the contributions of the pseudo labels to the self--training cross-entropy loss according to their likelihoods or, equivalently, the confidence of their belonging to the current cluster\footnote{The code is available at \url{https://github.com/qlilx/ICSR}.}.

\deleted{The contributions of the paper are as follows: }

\deleted{(1) A new image clustering technique that builds on and improves existing models by first ranking samples within the clusters based on the confidence of the samples belonging to their current clusters and then using the ranking to formulate a weighted cross-entropy loss to learn to improve clustering performances.
}

\deleted{(2) A method for computing the likelihood of samples belonging to their current clusters based on whether they are situated in densely populated neighborhoods and a scheme for weighting the ranked samples.}

\deleted[comment={deleted}]{(3) Extensive experimental results demonstrating that the new technique can improve the existing State--of--the--Art image clustering models and can perform well on a variety of datasets of remote--sensing images.}

\added[comment={added}]{The organization of this paper is as follows. In the next section, the related works are given. Section \ref{sec:3} illustrates the methods, including how to rank images (or pseudo labels) based on their reliability, how to weight the cross-entropy loss, and how to evaluate the performance of clustering. Experimental results and a discussion are given in Sections \ref{sec:4} and \ref{sec:5}, and a brief conclusion can be found in the last section. }

\section{Related Work}

\textbf{Representation learning.} 
Supervised representation learning is limited by the availability of manually labeled data. Various unsupervised representation learning methods have been proposed and have achieved outstanding performances. The methods of pretext tasks learn the representations of the images from the pre-designed tasks, and typical examples include inpainting patches~\cite{conf/cvpr/PathakKDDE16}, colorizing images~\cite{conf/eccv/ZhangIE16,conf/cvpr/LarssonMS17}, predicting the patch context~\cite{conf/iccv/DoerschGE15,conf/cvpr/MundhenkHC18}, 
solving jigsaw puzzles~\cite{conf/cvpr/NorooziVFP18}, and predicting rotations~\cite{conf/iclr/GidarisSK18}. 
Contrastive methods~\cite{henaff2019dataefficient,HintonETAL:06,conf/icml/LeeGRN09,conf/cvpr/TangSH12,JMLR:v11:vincent10a,conf/cvpr/RenL18,conf/cvpr/JenniF18,conf/nips/XieDDHN17,conf/nips/DonahueS19} 
use contrastive loss to make the representations of positive samples closer while negative ones further apart. 
The latent distributions in generating models~\cite{JMLR:v11:vincent10a,journals/jmlr/Baldi12,conf/iclr/DonahueKD17,conf/nips/DonahueS19} 
are not only much simpler than the input distribution, but also can capture semantic variation in the distributions of the input data. Besides these methods, integrating the clustering and optimization of neural networks also can learn good representations
~\cite{conf/eccv/CaronBJD18,conf/icml/HuangDGZ19,conf/iclr/AsanoRV20a,li2021designing}. In the past two years, self-supervised learning methods have been widely applied to the area of remote sensing and have reached remarkable performance \cite{9397864,rs13214255,9522871,li2022global,akiva2021self,ayush2021geography,rs12111868,rs14102425}.

\textbf{Image clustering.} Since direct clustering of images usually suffers from the curse of dimensionality problem, it is important to perform clustering using low-dimensional features or representations.
The clustering--based representation learning methods
~\cite{conf/eccv/CaronBJD18,conf/iclr/AsanoRV20a,li2021designing} can also directly perform clustering according to the features extracted from the neural network.
Other clustering methods by grouping features include DEC~\cite{xie2016unsupervised}, DAC~\cite{conf/iccv/ChangWMXP17}, and deep clustering~\cite{conf/iccv/CaronBMJ19}. Maximizing the mutual information between images and augmentations is another clustering method, e.g., IIC~\cite{conf/iccv/JiVH19,conf/icml/HuMTMS17}. 
 SCAN~\cite{conf/eccv/GansbekeVGPG20} takes two steps to implement image clustering: first, obtaining semantically meaningful features from a self-supervised task; second, using those features as a prior in a learnable clustering method. RUC~\cite{park2021improving} proposes a robust learning method to improve the pre-trained clustering models by SCAN and TSUC~\cite{han2020mitigating} via cleansing and refining labels. SPICE~\cite{niu2021spice} proposes two semantic-aware pseudo labeling algorithms, which provide accurate and
reliable self-supervision for~clustering.

\section{Method}
\label{sec:3}
A new method to Improve Clustering through Sample Ranking (ICSR) is introduced in this section. This method builds on the pretrained clustering models such as~\cite{conf/eccv/GansbekeVGPG20,niu2021spice}, which are trained based on the representations learned from self-supervised learning. ICSR ranks the samples in each cluster through a confidence criterion that can estimate the likelihoods of samples belonging to the current clusters.
The ranking of the samples is then used to weight their pseudo labels to form a modified cross-entropy loss. 

\subsection{Sample Ranking}\label{gradingimages}

Consider $K$ clusters of images where the clusters are formed by deep learning models, such as the models trained by the clustering--based representation learning method \cite{conf/iclr/AsanoRV20a,li2021designing}. In each cluster, assuming there is a dominant class, i.e., one class has the largest number of samples, we can regard samples from the dominant class as the signal and the others as noise. 
The first task of ICSR is to estimate the 
likelihood that an image is noise or not in a cluster.

Assume that the distance, such as the Euclidean distance (which can be replaced by any other functions of similarity such as cosine similarity or Kullback--Leibler divergence):
\bq\label{dis}
d(i,j) = |x_i - x_j|
\eq
in the feature space $X$ can indicate the similarity of two images $I_{(i)},\;I_{(j)}$, where $x_i,\;x_j$ are representations of the two images in the feature space $X$. For two arbitrary images $I_{(a)},\;I_{(b)}$ in the same cluster, find their $k$ nearest neighbors and 
compute the distances:
\bea\label{dk}
&&\{d(a,a_1),~d(a,a_2),\cdots,d(a,a_k)\} \nonumber \\
&&\{d(b,b_1),~d(b,b_2),\cdots,d(b,b_k)\}.
\eea
where $I_{(a_1)},I_{(a_2)},\ldots,I_{(a_k)}$ are the $k$ nearest neighbors of $I_{(a)}$ and $I_{(b_1)},I_{(b_2)},\ldots,I_{(b_k)}$ are the $k$ nearest neighbors of $I_{(b)}$. Based on these distances, if 
\bea\label{mean}
\text{mean}\left\{d(a,a_1),\cdots,d(a,a_k)\right\}> \text{mean}\left\{d(b,b_1),\cdots,d(b,b_k)\right\},
\eea
then it is reasonable to assume that $I_{(a)}$ has a higher probability to be noise than $I_{(b)}$.
Note that we assume there is a dominate class in each cluster, and the distance (\ref{dis}) could give the similarities of the samples; thus, the explanation of the criterion in (\ref{mean}) is obvious: an image that is not noise has more close neighbors (in a more densely populated region in the feature space).
As the mean value is susceptible to outliers, we also use the median value of the $k$ distances, i.e., if
\bea\label{median}
\text{median}\left\{d(a,a_1),\cdots,d(a,a_k)\right\}> \text{median}\left\{d(b,b_1),\cdots,d(b,b_k)\right\},
\eea
then we assume that $I_{(a)}$ is more likely to be noise than $I_{(b)}$. Again, the interpretation is that an image that is not noise has a higher chance to have closer neighbors.

For a given $k$ and an arbitrary image $I_{(i)}$, we can find the $k$ nearest neighbors of $I_{(i)}$, $I_{(i_1)},I_{(i_2)},\ldots,I_{(i_k)}$ and compute
\bea\label{criterion}
M = \text{mean}\left(d(i,i_1),~d(i,i_2),\cdots,d(i,i_k)\right) +\text{median}\left(d(i,i_1),~d(i,i_2),\cdots,d(i,i_k)\right).
\eea
$M$ then can be regarded as the score that indicates the 
likelihood of that image $I_{(i)}$ being noise: the higher the score, the higher the probability that it is noise. Using $M$ as a criterion to determine whether an image is the noise of a cluster is reasonable; however, $M$ is easily influenced by the choice of $k$.

\begin{figure}[H]
 \centering
 \includegraphics[width=16.8cm]{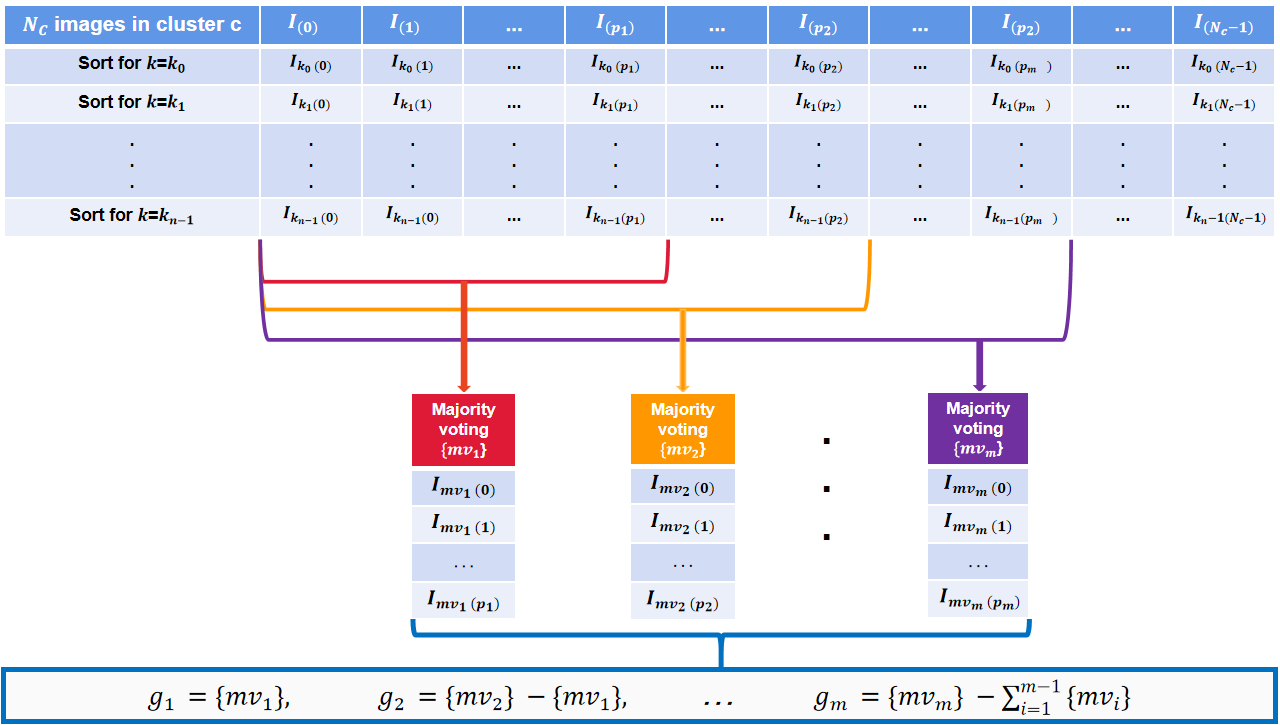}\\
 \caption{Schematic of the robust majority voting technique (also see Algorithm \ref{alg:algorithm}). The first row is the $N_c$ image samples in a cluster $c$. The following $n$ rows are samples sorted for $n$ different choices of $k$. Majority voting is then applied to group the samples based on the following procedure. Step 1: 
 Pick the first $p_1$ samples from each of the $n$ sorted lists and count how many times they appear.
 The top $p_1$ samples receiving the most votes (appeared most frequent) are kept to form Group 1, $g_1$. Step 2: Follow the same procedure as Step 1, but replace $p_1$ with $p_2$ ($p_2 > p_1$) to obtain the top $p_2$ samples receiving the most votes, then remove those samples already included in Group 1 to form Group 2, $g_2$. Step 3: Follow the same procedure as Step 1, but replace $p_1$ with $p_3$ ($p_3>p_2>p_1$) to obtain the top $p_3$ samples receiving the most votes, then remove those samples already included in previous groups ($g_1$ and $g_2$) to form Group 3, $g_3$. Subsequent groups can be formed following this pattern, and in total, we can obtain $m$ groups of such samples. It is not difficult to understand that samples in group $i$ will be less likely to be noise than samples in group $i+1$. Samples in the same cluster, but different groups are assigned the same pseudo labels, but differently weighted based on the likelihood of them belonging to noise.
 }\label{division}
\end{figure}

To reduce the effects of different selections of $k$, we introduce a robust majority voting technique, as illustrated in Figure \ref{division}. For each cluster and for an arbitrary allowed $k_i$, where $i = 0, 1, \cdots, n-1$, we can grade every sample in the cluster according to $M$ computed in (\ref{criterion}) and sort them according to the values of $M$ from low to high, i.e., from low to high likelihood that an image is noise in the cluster. In this way, we can obtain $n$ sorted sample lists. We then use majority voting to obtain $m$ groups of samples 
as {Algorithm} \ref{alg:algorithm}. 
It is easy to understand that samples in $g_i$ are less likely to be noise than samples in $g_{i+1}$. Based on this grouping, we modify the cross-entropy loss to improve the clustering performances.

\added[comment={added}]{In Algorithm \ref{alg:algorithm}, to compute $M$ in (\ref{criterion}) conveniently, an $N_c\times (N_c-1)$ matrix is constructed to store the distances among representations of images, where $N_c \approx N/k$ and $N,\;k$ are the image number and cluster number. Clearly, the space needed to store $N_c\times (N_c-1)$ is proportional to $N^2$, and thus, the space complexity of the algorithm is $O(N^2)$. Similarly, the temporal complexity of the algorithm is also $O(N^2)$ with respect to the image number. There is another factor that would affect the temporal complexity, that is the voting number $n$. For this variable, the temporal complexity of the algorithm is $O(n)$. }

\begin{algorithm}[H]
\caption{ Sample ranking and robust majority voting.}
\label{alg:algorithm}
\textbf{Input}: $N_c$ images in cluster $c$: $I_{(i)}, i=0,1, \cdots$, $N_c$\\
\textbf{Parameter}: $k_0$, $n$, $\hat k$, $m$, $p_1<p_2< \cdots<p_m$\\
\textbf{Output}: $m$ groups of images
\begin{algorithmic}[1] 
\STATE Select $n$ different $k$ $\{k_0, k_1, k_2, \cdots, k_{n-1}\}=\{k_0, k_0+\hat k,k_0 + 2\hat k, \cdots, k_0+(n-1)\hat k \}$
\STATE Compute $M(k_0),M(k_1),\cdots, M(k_{n-1}) $ for every image via Equation (\ref{criterion})
\STATE Sort images according to small $M$ $\rightarrow$ large $M$:\\
 $\{I_{k_0(0)},I_{k_0(1)},\cdots, I_{k_0(N_c-1)}\}$\\
$\{I_{k_1(0)},I_{k_1(1)},\cdots, I_{k_1(N_c-1)}\}$\\
$\cdots$\\
$\{I_{k_{n-1}(0)},I_{k_{n-1}(1)},\cdots, I_{k_{n-1}(N_c-1)}\}$\\ \label{sortlist}
\STATE $j=1$
\WHILE{$j\leq m$}
\STATE Count how many times $I_{(i)}$ appeared in the top $p_j$ of all the sorting lists in Step \ref{sortlist}.
\IF {$I_{(i)}$ in top $p_j$ most counted}
\IF {j==1}
\STATE Rank $I_{(i)}$ into group $g_1$.
\ELSE 
\STATE Rank $I_{(i)} \notin g_{j'<j}$ into group $g_j$
\ENDIF
\ELSE
\STATE Do nothing.
\ENDIF
\STATE j+=1
\ENDWHILE
\STATE \textbf{return} $g_1,g_2,\cdots, g_m$
\end{algorithmic}
\end{algorithm}

\subsection{Model Training}
\begin{figure}[H]
\centering
 \includegraphics[width=12.8cm]{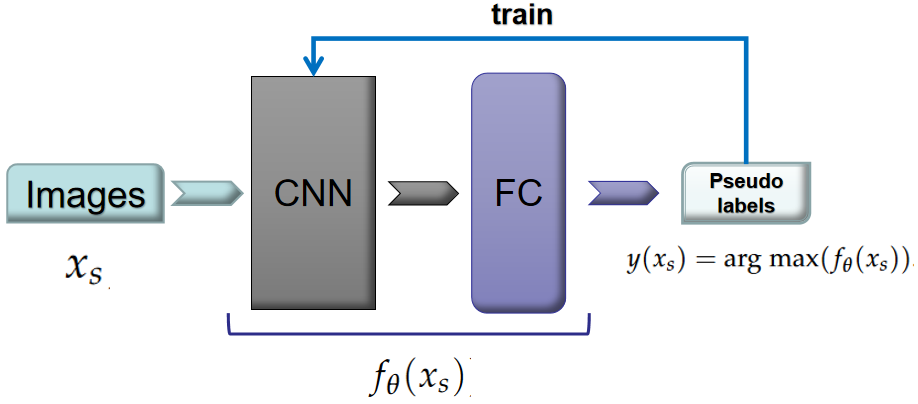}
 \caption{{Using the predictions of the network as pseudo labels that could be used to train the~network.}}
 \label{model}
\end{figure}

In this paper, we trained the model by standard cross-entropy loss:
\bq
L_{CE} = \frac{1}{S} \sum^{S}_{s = 1} \text{H} \left(y(x_s), f_\theta(x_s)\right).
\eq
where $f_\theta(x_s)$ is the clustering model, which is normally implemented as Figure \ref{model} in a convolutional neural network architecture such as ResNet or other popular architecture, and $y(x_s)$ is the pseudo label given by the model prediction as
\bq\label{pseudo_labels}
y(x_s) = \text{arg}\;\text{max}\left(f_\theta(x_s)\right).
\eq
\vspace{-22pt}

Based on the simple rationale that different augmentations of the same image should give the same prediction, i.e., the same pseudo label, we applied Label-Consistent Training (LCT) \cite{li2021designing}:
\bq
L_{LCT} = \frac{1}{S} \sum^{S}_{s = 1} \sum_{ij}\text{H} \left(y(\text{Aug}_i(x_s)), f_\theta(\text{Aug}_j(x_s))\right).
\eq
where $i, j$ represent different augmentations and the augmentations for creating pseudo labels are simpler. 
Minimizing the loss $L_{LCT}$ enables an input's different augmentations to give the same prediction, which plays the same role of consistency regularization~\cite{conf/nips/BachmanAP14} applied in FixMatch~\cite{sukhbaatar2014training}, RUC~\cite{conf/cvpr/PathakKDDE16}, and so on.

In Section \ref{gradingimages}, we propose a method to group the data according to their likelihood to be noise. For the $m$ groups of data, we should give different weights to their loss since the reliability of the data is different. Therefore, we modify the loss function as: 
\bea\label{loss_w}
L = \sum^m_i w_i L_{LCT}(x(g_i))
\eea
where the hyper-parameter $w_i$ is the weight of the loss function produced from the data of group $g_i$. The values of these parameters must satisfy the following relations:
\bq
w_1>w_2>w_3>\cdots>w_m
\eq
due to the reliability of the data in them decreasing from Group $1$ to group $m$. As the training proceeds, $w_2, w_3, \cdots w_m$ should gradually increase for two reasons: firstly, to explore the good weights for them; secondly, as accuracy increases, more samples are more reliable. In this paper, we adopted a strategy for computing the hyper-parameters $w_i$ as
\bq\label{wi}
w_i(t) = 1 - \left(\frac{i-1}{m}\right)^{(1+t)\beta_0},
\eq
where $i$ is the rank of the samples in group $g_i$ obtained from the ranking algorithm and $t$ is the training epoch index. $\beta_0$ is the (only) free parameter set manually, and we found setting it to between $0.01$ and $0.05$ works well (see Section \ref{ablation}, ablation study). As $\beta_0$ in (\ref{wi}) increases, $w_1$ will not be affected, while the values of $w_2, w_3,\cdots,w_m$ will increase, but never cross 1.

\subsection{Evaluation}

In this paper, we employ three approaches to evaluate our clustering: clustering Accuracy (ACC), Normalized Mutual Information (NMI) \cite{estevez2009normalized,amelio2015normalized}, and Adjusted Rand Index (ARI) \cite{rand1971objective, hubert1985comparing,vinh2010information}. ACC is the same as the accuracy of classification, but the labels given by the clustering are different from the ground truth. The clustering accuracy can be computed from the following equation:
\bq
\text{ACC} = \text{max}_m \frac{\sum_{i=1}^{n}\{y_i=m(c_i)\}}{n},
\eq
where $y_i$ is the ground truth, while $c_i$ is the clustering labels. $m$ is a mapping that maps clustering labels $c_i$ to the ground truth $y_i$. NMI evaluates clustering by measuring the mutual information between the ground truth labels and the clustering labels. It is an information theoretic metric normalized by the average of entropy of both the assignments of clustering and the ground truth labels as
\bq
\text{NMI}(A, B) = \frac{I(A, B)}{\frac{1}{2}\left[H(A)+H(B)\right]},
\eq
where $I(A,B)$ is the mutual information of $A$ and $B$ and $H(A)$ and $H(B)$ are the entropy of $A$ and $B$. The NMI can be implemented as sklearn.metrics.normalized\_mutual\_info\_score by Sklearn. Considering all pairs of samples and counting pairs with the same or different assignments in the predicted and true clusterings, the Rand Index (RI) can compute a similarity measure between two clusterings. The corrected-for-chance version of the Rand Index is the Adjusted Rand Index:
\bq
\text{ARI} = \frac{\text{RI} - \text{Expected(RI)}}{\text{max(RI)} - \text{Expected(RI)}},
\eq 
which can be implemented by Sklearn as sklearn.metrics.adjusted$\_$rand$\_$score.

\section{Experiments}
\label{sec:4}

\subsection{Comparison to the State--of--the--Art}

\subsubsection{Dataset}


To fairly compare our method with others, we need to conduct our experiments on the datasets used by other methods. In this paper, we chose four datasets that are usually used in image clustering: CIFAR--10/100, SLT-10, and ImageNet--10. The basic information of these four datasets is given in Table \ref{data}.

\renewcommand\arraystretch{0.8}
\begin{table}[H]
\caption{Basic information of CIFAR--10/100, SLT-10, and ImageNet--10. \label{data}}
\setlength{\tabcolsep}{5mm}{
\begin{tabularx}{\textwidth}{llllll}
\toprule
\textbf{Dataset} & \textbf{Pixel RES} & \textbf{Class $\#$} & \textbf{Training Image $\#$} & \textbf{Testing Image $\#$ } \\
\midrule 
CIFAR--10  & $32\times32$   & 10  & 50,000 & 10,000 \\
CIFAR--100  & $32\times32$   & 100/20 & 50,000 & 10,000 \\
STL--10   & $96\times96$   & 10  & 5000 & 8000 \\
ImageNet--10 & resized to $96\times96$   & 10  & 13,000 & 0 \\
\bottomrule
\end{tabularx}}
\end{table}

The CIFAR--10/100 datasets \cite{krizhevsky2009learning} consist of 60,000 color images, 50,000 training images and 10000 testing images. For CIFAR--10, there are 10 classes, including airplane, automobile, bird, cat, deer, dog, frog, horse, ship, and truck. These classes are completely mutually exclusive, for example ``automobile'' includes SUVs, sedans, and things of that sort, while ``truck'' only includes big trucks. 

For CIFAR--100, there are only 500 training images and 100 testing images per class. The 100 classes are grouped into 20 superclasses and, thus, each image has a ``coarse'' label and a ``fine'' label. The superclasses include aquatic mammals, fish, flowers, food containers, fruits and vegetables, household electrical devices, insects, and so on. Each superclass contains 5 classes, such as ``aquatic mammals'' includes beaver, dolphin, otter, seal, and whale and ``flowers'' includes orchids, poppies, roses, sunflowers, and tulips.

SLT-10 \cite{coates2011analysis} is a dataset for developing unsupervised feature learning, deep learning, and self-taught learning algorithms. It is a modified version of CIFAR--10. There are fewer labeled training images and a large amount of unlabeled images that are from a similar, but different distribution of the labeled data. However, in this paper, we only used the labeled images, but without using their labels in training. The 10 classes of images in the this dataset include images of airplanes, birds, cars, cats, deer, dogs, horses, monkeys, ships, and trucks. In this paper, the training and evaluations are on all labeled data, 5000 training images and 8000 testing images. 

ImageNet--10 dataset is a subset of ILSVRC2012 \cite{russakovsky2015imagenet} that contains around 1.28 million images with 1000 classes. The pixel resolutions of the images in ILSVRC2012 are different, and in the experiments, we resized all images to $96\times 96$ pixels. The 10 classes used in this paper contain penguin, dog, leopard, airplane, airship, freighter, football, sports car, truck, and~orange.

\subsubsection{Results}

To evaluate the effectiveness of the proposed Image Clustering though Sample Ranking (ICSR) method, we applied it to models pretrained by the State--of--the--Art image clustering and representation learning methods with the aim of further improving their performances. It is important to note that even though the experimental data contain labels, the labels were never used in training.

We tested our method on the three State--of--the--Art clustering models, SCAN~\cite{conf/eccv/GansbekeVGPG20}, RUC~\cite{park2021improving}, and SPICE~\cite{niu2021spice}, on widely used standard datasets, CIFAR--10/100, STL--10, and ImageNet--10. \deleted[comment={deleted}]{There are 50,000 training and 10,000 testing images of 32$\times$32 pixels in CIFAR--10/100 (10/100 classes). In all experiments of CIFAR--10/100 we only use the training samples to train the models and evaluate both on training and testing images. For CIFAR--100, as the previous work, we grouping the images into 20 clusters, which could correspond to the coarse labels of CIFAR--100. For STL--10 (10 classes), there are 13000 labeled images of $96\times96$ pixels: 5000 training images and 8000 testing images. In the training of models on STL--10, we use all 13000 samples (but without using their labels). There are also 13000 images in ImageNet--10 (10 classes), and we resize all images of ImageNet to $96\times96$ pixels before training. } \added[comment={added}]{SCAN is a two-step method where visual representation learning and clustering are decoupled: firstly, employing a self-supervised learning (MoCo \cite{conf/cvpr/He0WXG20} or SimCLR \cite{journals/corr/abs-2002-05709}) to learn semantically meaningful features; secondly, using the obtained features as a prior to perform clustering (classify each image and its nearest neighbors together). Additionally, a self--training technique was used to improve clustering performance. self--training is training a model using the pseudo labels that are given by Equation~(\ref{pseudo_labels}). RUC is a method of robust learning, which is very similar to ours, and the difference is mainly on how to treat pseudo labels. In RUC, three strategies are proposed to deal with pseudo labels: first, filtering out samples with low prediction confidence; second, detecting cleaned samples by checking if the given sample has the same label with the top $k$ nearest neighbor; third, combining the above two strategies. In SPICE, there are 3 training stages: first, feature model training; second, cluster head training; third, feature model and cluster head joint training. SPICE designs a prototype pseudo labeling method where top confident samples are chosen to estimate the prototypes for each cluster, and their indices are then assigned to their nearest neighbors as the pseudo labels.}

We used ResNet-18 for CIFAR--10/100 and STL--10, while WRN-37-2 for ImageNet--10, as previous works did. The data augmentation used is similar to~\cite{conf/eccv/GansbekeVGPG20,park2021improving}. 
To test the effectiveness of our method for other network architectures, we also conducted the experiments on AlexNet for CIFAR--10 and STL--10, where we established the clustering models through the clustering--based representation learning method~\cite{li2021designing}. We trained all the models with 240 epochs with $\beta_0=0.02$, a learning rate of 0.005, a batch size of 128, and an SGD momentum of 0.9. We evaluated all models by both clustering Accuracy (ACC), Normalized Mutual Information (NMI), and the Adjusted Rand Index (ARI).

For ranking samples or the pseudo labels, we used the Euclidean distance of the final layer of the network to compute $M$ in Equation~(\ref{criterion}). Choosing the distance of the final layer but the CNN layers was due to most models here experiencing self--training, and thus, the final layer is representative enough and has lower dimensions. $k_0$ in Algorithm \ref{alg:algorithm} takes 1/3 of the number of samples in the clusters, and the step $\hat k$ takes 5. The largest $k_{n-1}$ was set to close to 2/3 of the number of samples in the clusters. We ranked samples into 5 groups. In fact, the case of setting less than 5 groups was roughly included due to, in the early training epoch, $w_2,\;w_3,\;w_4,\;w_5$ are extremely small (see the ablation study). Setting more group numbers can help a little, but needs much more computing resources. $p_1,\;p_2\cdots p_5$ in Algorithm \ref{alg:algorithm} take $15\%$, $35\%$, $55\%$, $75\%$, $95\%$ of the samples in each cluster and increase $1\%$ for every 50~epochs. 

The performances of our method are shown in Table \ref{com}. We can see that applying our ICSR method to the pretrained models from SACN, RUC, and SPICE, the clustering performances markedly improved. 

For CIFAR--10, ICSR can improve all these clustering models. 
For example, the ACC, NMI, and ARI performances of the SCAN model improved by $4.7\%$, $5.6\%$, and $8.3\%$, respectively. For the best model on CIFAR--10, SPICE, which already achieved a remarkably high accuracy of $92.6\%$, ICRS can still achieve a noticeable improvement of $2.1\%$. RUC is also a method proposed to improve image clustering. 
Results in Table \ref{com} show that models trained by SCAN + RUC can be further improved by ICSR. 

Although these State--of--the--Art clustering models have quite different performances on STL--10, ranging from $81.7\%$ to $92.0\%$, our method works very well consistently on them, achieving significant performance gains ranging from $6.0\%$ to $13.6\%$. 
It is worth noting that the best clustering accuracy on SLT-10 was improved by our method to $98\%$, close to a perfect $100\%$.

For ImageNet--10, SPICE already reached a very high level of performance of $95.9\%$. Still, applying the new ICSR can further improve the clustering accuracy by $2.2\%$. For CIFAR--100, there are 20 classes. Although in a few clusters, there is no dominant class, our method also can improve the performances of the three clustering models. 

\renewcommand\arraystretch{0.8}
\begin{table} [H]
\caption{Evaluation of image clustering performances. Building on three State--of--the--Art clustering models, SCAN, RUC, and SPICE, ICSR can significantly improve these models, achieving performance gains ranging from $2.0\%$ to $13.6\%$. For the case where the performance is already very high $95.9\%$, ICSR can still improve it further to $98.1\%$, very close to a perfect $100\%$. The best results are highlighted by the bold. \label{com}}
\setlength{\tabcolsep}{5mm}{
\begin{tabularx}{\textwidth}{lllllll}
\toprule
\textbf{Method} & \multicolumn{3} {c} {\textbf{CIFAR--10}} & \multicolumn{3} {c} {\textbf{CIFAR--100--20}} \\
\textbf{Evaluation} & \textbf{ACC} & \textbf{NMI} & \textbf{ARI} & \textbf{ACC} & \textbf{NMI} & \textbf{ARI} \\
\midrule
k-means~\cite{macqueen1967some}  & 22.9 & 8.7 & 4.9 & 13.0 & 8.4 & 2.8  \\
SC~\cite{ng2002spectral}    & 24.7 & 10.3 & 8.5 & 13.6 & 9.0 & 2.2  \\
AC~\cite{franti2006fast}    & 22.8 & 10.5 & 6.5 & 13.8 & 9.8 & 3.4  \\
GAN~\cite{radford2015unsupervised} & 31.5 & 26.5 & 17.6 & 15.1 & 12.0 & 4.5  \\
DEC~\cite{xie2016unsupervised}  & 30.1 & 25.6 & 16.1 & 18.5 & 13.6 & 5.0  \\
DAC~\cite{zhou2018deep}     & 55.2 & 39.6 & 30.9 & 23.8 & 18.5 & 8.8  \\
DeepCluster~\cite{conf/eccv/CaronBJD18}  & 37.4 & - & - & 18.9 & - & -  \\
DDC~\cite{chang2019deep}    & 52.4 & 42.4 & 32.9 & - & - & -  \\
IIC~\cite{conf/iccv/JiVH19}   & 61.7 & - & - & 25.7 & - & -  \\
TSUC~\cite{han2020mitigating}   & 61.7 & - & - & 35.5 & - & -  \\
SCAN~\cite{conf/eccv/GansbekeVGPG20} & 88.7 & 80.4 & 78.0 & 50.6 & 47.5 & 32.9 \\
SCAN+RUC~\cite{park2021improving} & 90.3 & 83.2 & 80.9 & 54.3 & 55.1 & 38.7  \\
SPICE~\cite{niu2021spice}    & 92.6 & 85.8 & 84.3 & 53.8 & 56.7 & 38.7 \\
\midrule
SCAN + ICSR        & 93.4 & 86.0 & 86.3 & 54.4 & 51.7 & 35.9 \\
SCAN + RUC + ICSR      & 94.0 & 87.2 & 87.3 & 57.3 & 58.5 & 41.2 \\
SPICE + ICSR       & \textbf{94.7} & \textbf{88.6} & \textbf{88.9} & \textbf{58.8} & \textbf{59.6} & \textbf{42.1} \\
\midrule
 & \multicolumn{3} {c} {STL--10} & \multicolumn{3} {c} {ImageNet--10} \\
\midrule
k-means~\cite{macqueen1967some}  & 19.2 & 12.5 & 6.1 & 24.1 & 11.9 & 5.7 \\
SC~\cite{ng2002spectral}    & 15.9 & 9.8 & 4.8 & 27.4 & 15.1 & 7.6 \\
AC~\cite{franti2006fast}    & 33.2 & 23.9 & 14.0 & 24.2 & 13.9 & 6.7 \\
GAN~\cite{radford2015unsupervised} & 29.8 & 21.0 & 13.9 & 34.6 & 22.5 & 15.7 \\
DEC~\cite{xie2016unsupervised}  & 35.9 & 27.6 & 18.6 & 38.1 & 28.2 & 20.3 \\
DAC~\cite{zhou2018deep}     & 47.0 & 36.6 & 25.7 & 52.7 & 39.4 & 30.2 \\
DeepCluster~\cite{conf/eccv/CaronBJD18}  & 33.4 & - & - & - & - & -  \\
DDC~\cite{chang2019deep}    & 48.9 & 37.1 & 26.7 & 57.7 & 43.3 & 34.5 \\
IIC~\cite{conf/iccv/JiVH19}   & 61.0 & - & - & - & -  & -  \\
TSUC~\cite{han2020mitigating}   & 62.0 & - & - & - & -  & -  \\
SCAN~\cite{conf/eccv/GansbekeVGPG20} & 81.4 & 69.8 & 64.6 & - & -  & -  \\
SCAN+RUC~\cite{park2021improving} & 86.7 & 77.8 & 74.2 & - & -  & -  \\
SPICE~\cite{niu2021spice}    & 92.0 & 85.2 & 83.6 & 95.9 & 90.2 & 91.2 \\
\midrule
SCAN + ICSR        & 95.0 & 89.4 & 89.5 & - & - & -  \\
SCAN + RUC + ICSR      & 94.8 & 89.0 & 89.2 & - & - & -  \\
SPICE + ICSR       & \textbf{98.0} & \textbf{95.1} & \textbf{95.8} & \textbf{98.1} & \textbf{94.8} & \textbf{95.7} \\
\bottomrule
\end{tabularx}}
\end{table}

Table \ref{ucl} gives the evaluation results on testing images, which were never involved in the training. The accuracy of this evaluation can demonstrate the models' capability of generalization. 

For CIFAR--10, our method can boost the performance of all three clustering models. Although SPICE achieved a very high performance, very close to the accuracy of supervised learning, $93.8\%$, our ICSR method can further close the gap between supervised and unsupervised learning by $1.0\%$. 

For CIFAR--100--20, for the model pretrained by SCAN + RUC, although our method improved its clustering performances on the training data (see Table \ref{com}), it did not on the testing data. However, for the other two pretrained models, the accuracies were improved by our method. For the model pretrained by SPICE, although SPICE + ICSR improved the clustering accuracy, it did not improve the NMI and ARI. 

\renewcommand\arraystretch{0.8}
\begin{table}[H]
\caption{{Unsupervised image classification on CIFAR--10. The model is trained on the training images without using their labels (unsupervised) and classification accuracy evaluated on the testing images. The best results are highlighted by the bold.} \label{ucl}}
\setlength{\tabcolsep}{5mm}{
\begin{tabularx}{\textwidth}{lllllll}
\toprule
\textbf{Method} & \multicolumn{3} {c} {\textbf{CIFAR--10}} & \multicolumn{3} {c} {\textbf{CIFAR--100--20}} \\
\textbf{Evaluation} & \textbf{ACC}& \textbf{NMI} & \textbf{ARI} & \textbf{ACC} & \textbf{NMI} & \textbf{ARI}  \\
\midrule
supervised & 93.8 & - & - & 80.0 & - & -  \\
SCAN   & 88.3 & 79.7 & 77.2 & 50.7 & 48.6 & 33.3   \\
SCAN + RUC & 89.1 & 81.5 & 78.7 & 53.4 & 54.9 & 37.8  \\
SPICE   & 91.8 & 85.0 & 83.6 & 53.5 & \textbf{56.5} & \textbf{40.4}   \\
\midrule
SCAN + ICSR  & 90.5 & 80.8 & 80.6 & 51.6 & 48.0 & 35.9   \\
SCAN + RUC + ICSR  & 91.0 & 81.8 & 81.6 & 52.5 & 50.3 & 34.7 \\
SPICE + ICSR & \textbf{92.8} & \textbf{85.1} & \textbf{85.1} & \textbf{54.8} & 52.6 & 36.4\\
\bottomrule
\end{tabularx}}
\end{table}

In Tables \ref{com} and \ref{ucl} above, we tested our method on the models trained as a classifier via pseudo labels. To demonstrate that our method can also be applied to other models trained differently, we considered two other kinds of models. First, we considered models by SCAN, but without self-labeling. From Table \ref{scan*}, we see that on the model without the self-labeling steps, our method still had good performances, especially on the SCAN$^*$ models trained on STL--10. There was a $15.9\%$ improvement, which is even better than building on models pretrained from SCAN and RUC.

Secondly, we considered the models trained by clustering--based representation learning~\cite{li2021designing}. \added[comment={added}]{The method \cite{li2021designing} is proposed to learn visual representation through clustering. Its goal is visual representation, while clustering is the means to achieve the goal. To make the visual representations as discriminative as possible, \cite{li2021designing} evenly distributes images among clusters by translating the final layer of the network. Although \cite{li2021designing} is not a method for clustering, it could be used as the pretrained model for our method.}

\begin{table}[H]

\caption{SCAN* is the model from SCAN~\cite{conf/eccv/GansbekeVGPG20} without the self-labeling steps. It is a clustering model based on features extracted from unsupervised representation learning. \label{scan*}}
\setlength{\tabcolsep}{5mm}{
\begin{tabularx}{\textwidth}{lllllll}
\toprule
\textbf{Method} & \multicolumn{3}{c}{\textbf{CIFAR--10}} & \multicolumn{3}{c}{\textbf{STL--10}} \\
\textbf{Evaluation}    & \textbf{ACC} & \textbf{NMI} & \textbf{ARI} & \textbf{ACC} & \textbf{NMI} & \textbf{ARI}  \\
\midrule
SCAN$^*$     & 82.2 & 72.1 & 67.7 & 79.2 & 67.1 & 61.8 \\
SCAN$^*$ + ICRS     & 90.3 & 82.9 & 81.3 & 95.1 & 89.4 & 89.6 \\
\midrule
Improved     & 8.1 & 10.8 & 12.6 & 15.9 & 22.3 & 27.8 \\
\bottomrule
\end{tabularx}}
\end{table}

In this experiment, we trained the models on AlexNet and then applied the ICSR method to improve the image clustering. Table \ref{lct} demonstrates that the ICSR method can be applied to improve models from the clustering--based representation learning method, and it also works well on AlexNet. 

\begin{table}[H]
\caption{{ICSR improves} clustering performances on models trained by the clustering--based representation learning method. LCT is trained by representation learning method~\cite{li2021designing}, and the backbone of the model is AlexNet. \label{lct}}
\setlength{\tabcolsep}{5mm}{
\begin{tabularx}{\textwidth}{lllllll}
\toprule
\textbf{Method} & \multicolumn{3} {c} {\textbf{CIFAR--10}} & \multicolumn{3} {c} {\textbf{STL--10}} \\
\textbf{Evaluation}    & \textbf{ACC} & \textbf{NMI} & \textbf{ARI} & \textbf{ACC} & \textbf{NMI} & \textbf{ARI}  \\
\midrule
LCT      & 83.2 & 73.1 & 69.3 & 53.3 & 47.8 & 35.5 \\
LCT + ICRS    & 90.8 & 82.7 & 81.7 & 73.7 & 66.1 & 58.8 \\
\midrule
Improved     & 7.6 & 9.6 & 12.4 & 20.4 & 18.3 & 23.3 \\
\bottomrule
\end{tabularx}}

\end{table}

\subsection{Applied to remote--sensing images}

\subsubsection{Dataset}

 In this paper, we consider 7 datasets of remote--sensing images, as shown in Table \ref{dataset}. 

\begin{table}[H]
\caption{Datasets of remote--sensing images. \label{dataset}}
\setlength{\tabcolsep}{3.5mm}{
\begin{tabularx}{\textwidth}{lllllll}
\toprule
\textbf{Dataset} & \textbf{Pixel RES} & \textbf{Spatial RES} & \textbf{Class $\#$} & \textbf{Image $\#$} & \textbf{Source} \\
\midrule 
HMHR     & $256\times256$ & 2.39 m   & 5 & 533 & GoogleEarth \\
EuroSAT     & $64\times64$ & 10 m   & 10 & 27000 & Sentinel2 \\
SIRI WHU    & $200\times200$ & 2 m   & 12 & 2400 & GoogleEarth \\
UCMerced  & $256\times256$ & 0.3 m   & 21 & 2100 & USGS National Map \\
AID      & $600\times600$ & 0.5--8 m  & 30 & 10,000 & GoogleEarth \\
PatternNet   & $256\times256$ & 0.062--4.693 m & 38 & 30,400 & GoogleMap \\
NWPU   & $256\times256$ & 0.2--30 m  & 45 & 31,500 & GoogleEarth \\
 \bottomrule
\end{tabularx}}

\end{table}
The How--to--make--high--resolution--remote--sensing--image--dataset (HMHR) is made by Google map through LocalSpece Viewer. There are only 533 images in this dataset, which contains 5 classes: building, farmland, greenbelt, wasteland, water. 

EuroSAT \cite{helber2019eurosat,helber2018introducing} is a dataset for land use and land cover classification. It consists of 10 classes with 27000 remote--sensing images of annual crop, forest, highway, river, sea/lake, and so on. 

The dataset of SIRI WHU Google \cite{ma2015adaptive} is a dataset that includes 12 classes: agriculture, commercial, harbor, idle land, industrial, meadow, overpass, park, pond, river, water, residential. 

UCMerced LandUse \cite{yang2010bag} gives the images that were manually extracted from large images from the USGS National Map Urban Area Imagery collection for various urban areas around the country. There are 21 classes in this dataset, including airplane, beach, buildings, freeway, golf course, tennis court, and so on, and there are 100 images for each~class. 

AID \cite{xia2017aid} is an aerial image dataset with high resolution, 600 $\times$ 600 pixels, which is collected from Google Earth imagery. This dataset is not evenly distributed, and in each class, there are about 200 to 400 images. The scene classes contains bare land, baseball field, beach, bridge, center, church, dense residential, desert, forest, meadow, medium residential, mountain, park, parking, playground, port, railway station, resort, school, and so on. 

PatternNet \cite{zhou2018patternnet} is a remote sensing dataset collected for remote sensing image retrieval from Google Earth imagery or via the Google Map API for some U.S. cities. The 38 classes include airplane, cemetery, dense residential, forest, oil gas field, overpass, nursing home, parking lot, railway, closed road chaparral, bridge, and so on. 

NWPU RESISC45 \cite{cheng2017remote} is made by Northwestern Polytechnical University (NWPU), which is available for REmote Sensing Image Scene Classification (RESISC). The 45 scene classes include baseball diamond, basketball court, bridge, chaparral, church, circular-farmland, cloud, commercial area, dense residential, desert, intersection, island, lake, meadow, medium residential, mobile home park, wetland, and so on.

\subsubsection{Results}

In this part, we apply our method to the datasets introduced above. We first performed a clustering--based representation learning method \cite{li2021designing} to learn the representations of the images and then, based on these representations, to perform the clustering of remote--sensing images.

For the very small datasets, HMHR, UCMerced LandUse, and SIRI WHU Google, to prevent the network from overfitting, we used the strong augmentations as done in \cite{conf/eccv/GansbekeVGPG20} and the pretrained model that is trained on another remote sensing image dataset, i.e., PatternNet, by an unsupervised representation learning method \cite{li2021designing}. In the ranking process, we used the Euclidean distance of the last CNN layer, which is different from the experiments in the previous part in which the distance was computed from the final layer of the network. This is because, in this part, we establish all pretrained models from a representation learning method whose goal is not clustering. For all datasets of remote sensing, we used ResNet-18 as the backbone of the network, and all images were resized to $256\times256$ and then cropped to $224\times224$. The total training epochs were 240, which was divided into two sets of 120 epochs, and for the first 120 epochs, $p_1,\;p_2\cdots p_5$ in Algorithm \ref{alg:algorithm} take $15\%$, $35\%$, $55\%$, $75\%$, $95\%$ of the samples in each cluster, while for the second 120 epochs, $p_1,\;p_2\cdots p_5$ take $17.5\%$, $37.5\%$, $57.5\%$, $77.5\%$, $97.5\%$ of the samples in each cluster. For each 120 training epochs, $\beta_0 = 0.03$, and the learning rate initially was 0.03 and dropped to 0.003 at Epoch 90.

\begin{table}[H]

\caption{The clustering performances by our method for 7 datasets of remote--sensing images.\label{results}}
\setlength{\tabcolsep}{5mm}{
\begin{tabularx}{\textwidth}{lllll}

\textbf{Dataset} & \textbf{evaluation} & \textbf{LCT} & \textbf{LCT + ICSR} & \textbf{Improved} \\
\midrule 
     & ACC & 86.3 & 89.1 & 2.8 \\
HMHR    & NMI & 68.3 & 72.9 & 4.6 \\
     & ARI & 69.9 & 75.0 & 5.1 \\
\midrule
     & ACC & 90.0 & 94.8 & 4.8 \\
EuroSAT   & NMI & 84.4 & 89.7 & 5.3 \\
     & ARI & 80.3 & 89.3 & 9.0 \\
\midrule
     & ACC & 66.2 & 90.7 & 24.5 \\
SIRI WHU   & NMI & 66.8 & 86.1 & 19.3 \\
     & ARI & 53.3 & 82.0 & 28.7 \\
\midrule
     & ACC & 72.9 & 84.9 & 12.0 \\
UCMerced   & NMI & 78.3 & 86.5 & 8.2 \\
     & ARI & 62.3 & 77.5 & 15.2 \\
\midrule
     & ACC & 71.0 & 77.9 & 6.9 \\
AID    & NMI & 76.3 & 80.0 & 3.7 \\
     & ARI & 60.2 & 67.6 & 7.4 \\
\midrule
     & ACC & 90.9 & 97.1 & 6.2 \\
PatternNet  & NMI & 94.2 & 97.3 & 3.1 \\
     & ARI & 87.8 & 95.1 & 7.3 \\
\midrule
     & ACC & 76.1 & 85.4 & 9.3 \\
NWPU    & NMI & 80.5 & 84.8 & 4.3 \\
     & ARI & 66.9 & 75.4 & 8.5 \\
\bottomrule
\end{tabularx}}

\end{table}

Although these datasets are very different in terms of the source, size, spatial resolution, sample number, and class number (see Table \ref{dataset}), our method can work well on both them (see Table \ref{results}). For EuroSAT and PatternNet, the accuracies improved to $94.8\%$ and $97.1\%$ without using any labels, and for SIRI WHU Google, the accuracy improved $24.1\%$. For the very small datasets, such as HMHR and UCMerced LandUse, due to the samples being very limited, good ranking was difficult. However, our method still worked well on these datasets (the performance of ranking images in UCMerced LandUse can be found in the next part and \textbf{Appendix A, B}). 
For datasets AID, PatternNet, and NWPU RESISC45, the class numbers were 30, 38, and 45, which are large numbers for the clustering task. However, our method still worked on them. For AID, although the distribution of samples is not uniform and there are 200-400 images in each class, our method was still effective on it.

\section{Discussion}
\label{sec:5}
\subsection{Performance of Sample Ranking Algorithm}
\begin{figure}[H] 
\centering
\subfigure {
\includegraphics[width=0.47\columnwidth]{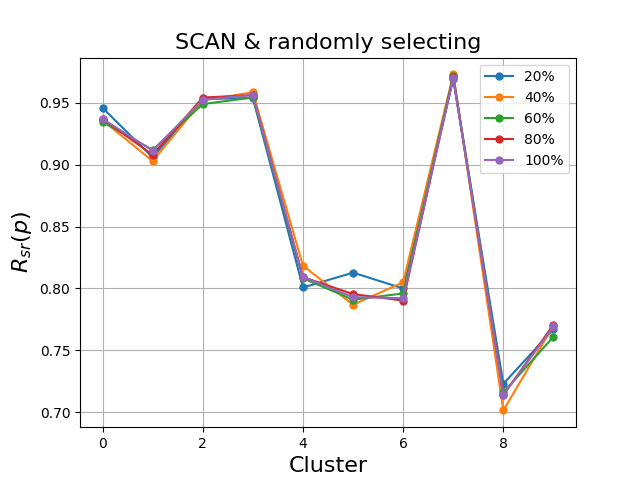}
}
\subfigure {
\includegraphics[width=0.47\columnwidth]{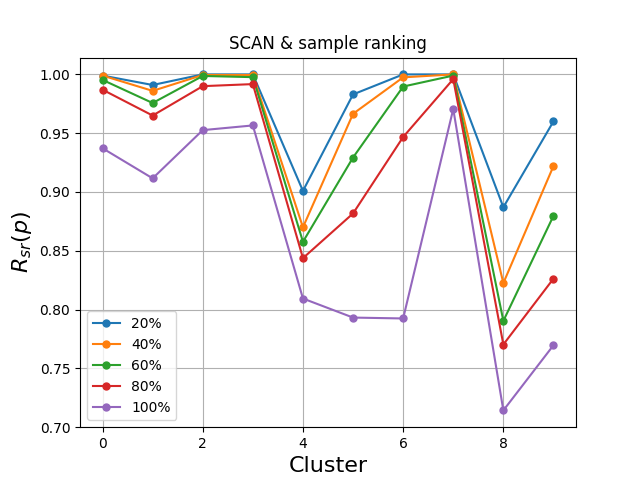}
}

\subfigure {
\includegraphics[width=0.47\columnwidth]{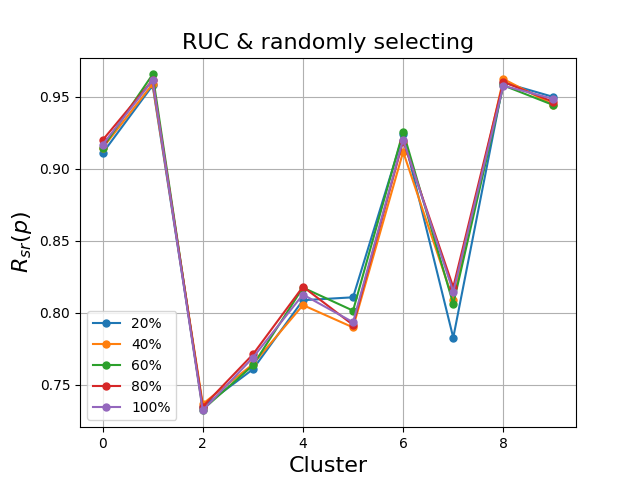}
}
\subfigure {
\includegraphics[width=0.47\columnwidth]{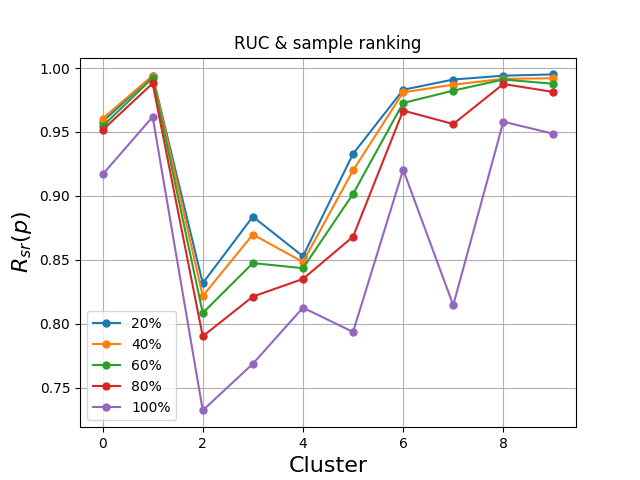}
}

\subfigure {
\includegraphics[width=0.47\columnwidth]{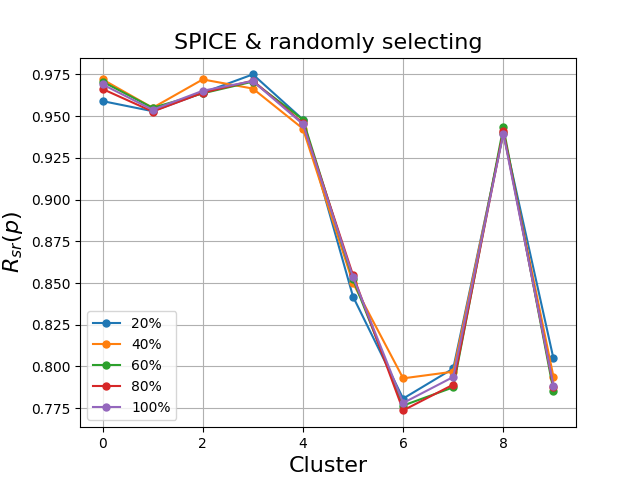}
}
\subfigure {
\includegraphics[width=0.47\columnwidth]{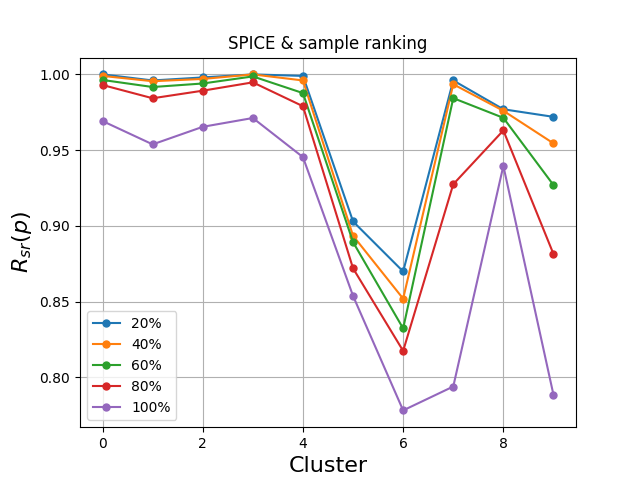}
}
\caption{Sample ranking performances of {Algorithm \ref{alg:algorithm}} (see also Figure \ref{division}). First column, $R_{sr}(p)$ without applying our ranking algorithm and randomly selecting $p=20\%,\; 40\%$, $\cdots$, $80\%$ of samples in the clusters formed by 3 pretrained clustering models for CIFAR--10. Second column, after applying our ranking algorithm to rank the samples in the clusters formed by the same 3 pretrained clustering model, $R_{sr}(p)$ for $p=20\%, 40\%$, $\cdots$, $80\%$.
}
\label{ranking}
\end{figure}

The core of the ICSR technique is the sample ranking and robust majority voting algorithm: Figure \ref{division} and {Algorithm \ref{alg:algorithm}}. ICSR assumes that each cluster has a dominant class, i.e., one class has more samples than other classes, and we call samples from the dominant class as the signal and other samples noise. The objective of the ranking algorithm is to place signal samples at a higher rank than the noise samples. The more signal samples are placed at a higher rank, the better the performances. We define the percentage of signal samples in the top-ranked $p\%$ of all samples in a cluster as the ranking success rate, $R_{sr}(p)$: 

\bq\label{rsr}
R_{sr}(p) = \frac{S(p)}{S(p)+N(p)} 
\eq
where $S(p)$ is the number of signal samples and $N(p)$ is the number of noise samples at the top $p\%$ of the ranked samples within a cluster. A good ranking performance should have the following characteristics. For a given $p$, the larger $R_{sr}(p)$ is, the better. For a given cluster and $p_i<p_j$, $R_{sr}(p_i)>R_{sr}(p_j)$. We illustrate the performance of ICSR's ranking algorithm in Figure \ref{ranking}. The graphs in the first column are consistent with the statistics, i.e., the percentage of signals in the sub-clusters is close to that of the whole cluster. The graphs in the second column demonstrate the effectiveness of our method, where in all three cases, the ranking success rate $R_{sr}(p)$ of the top $20\%$ is generally higher than that of the top $40\%$; $R_{sr}(p)$ of the top $40\%$ is higher than that of the top $60\%$, and so on. 
\vspace{-12PT}

\begin{figure}[H]
\centering
 \includegraphics[width=12.8cm]{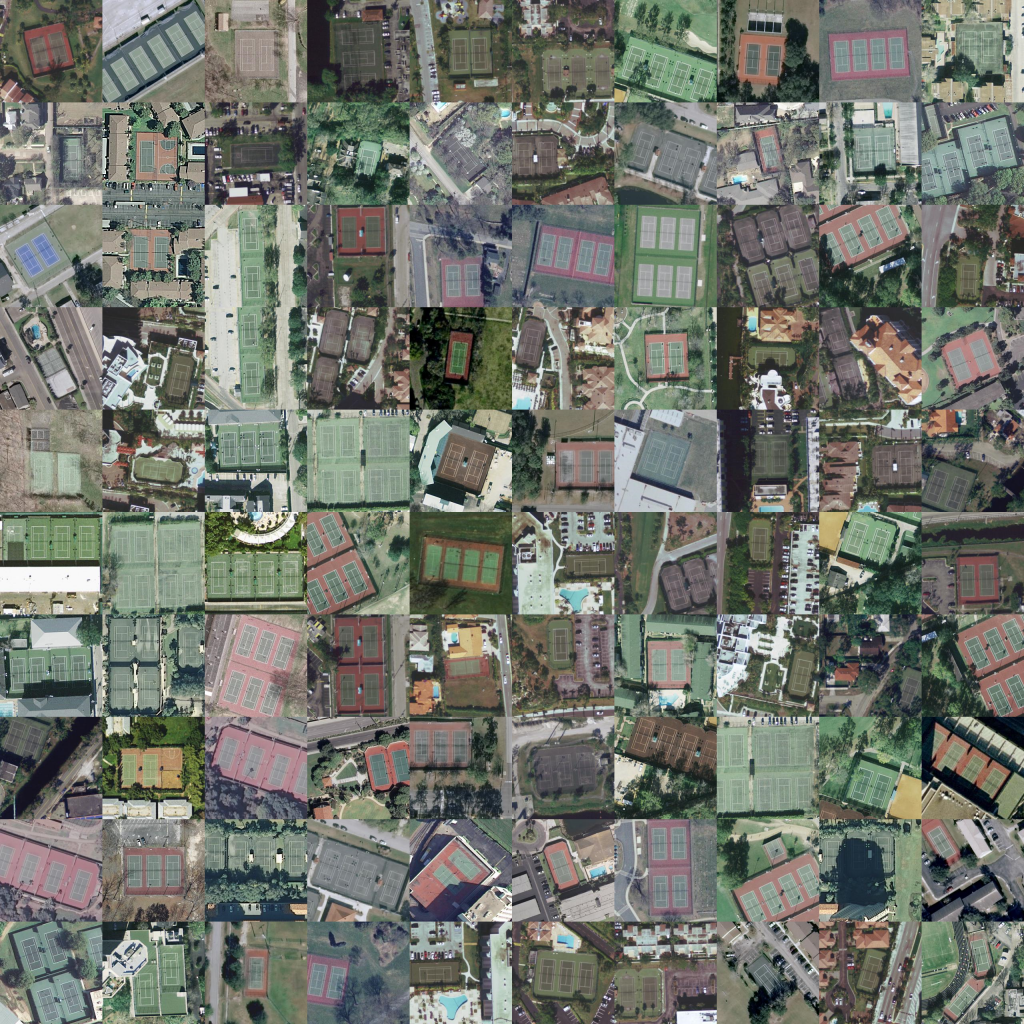}
 \caption{Ground truth for remote--sensing images of tennis courts in UCMerced LandUse. There are 100 images of tennis courts in total.}
 \label{tenis_groundtruth}
\end{figure}

To visualize the effectiveness of our ranking algorithm in remote--sensing images, we took dataset UCMerced LandUse as an example to illustrate it. Let us focus on one class of remote--sensing images, for example the tennis court. In the dataset, all images of tennis court are given in Figure \ref{tenis_groundtruth}.

\begin{figure}[H]
\centering
 \includegraphics[width=12.8cm]{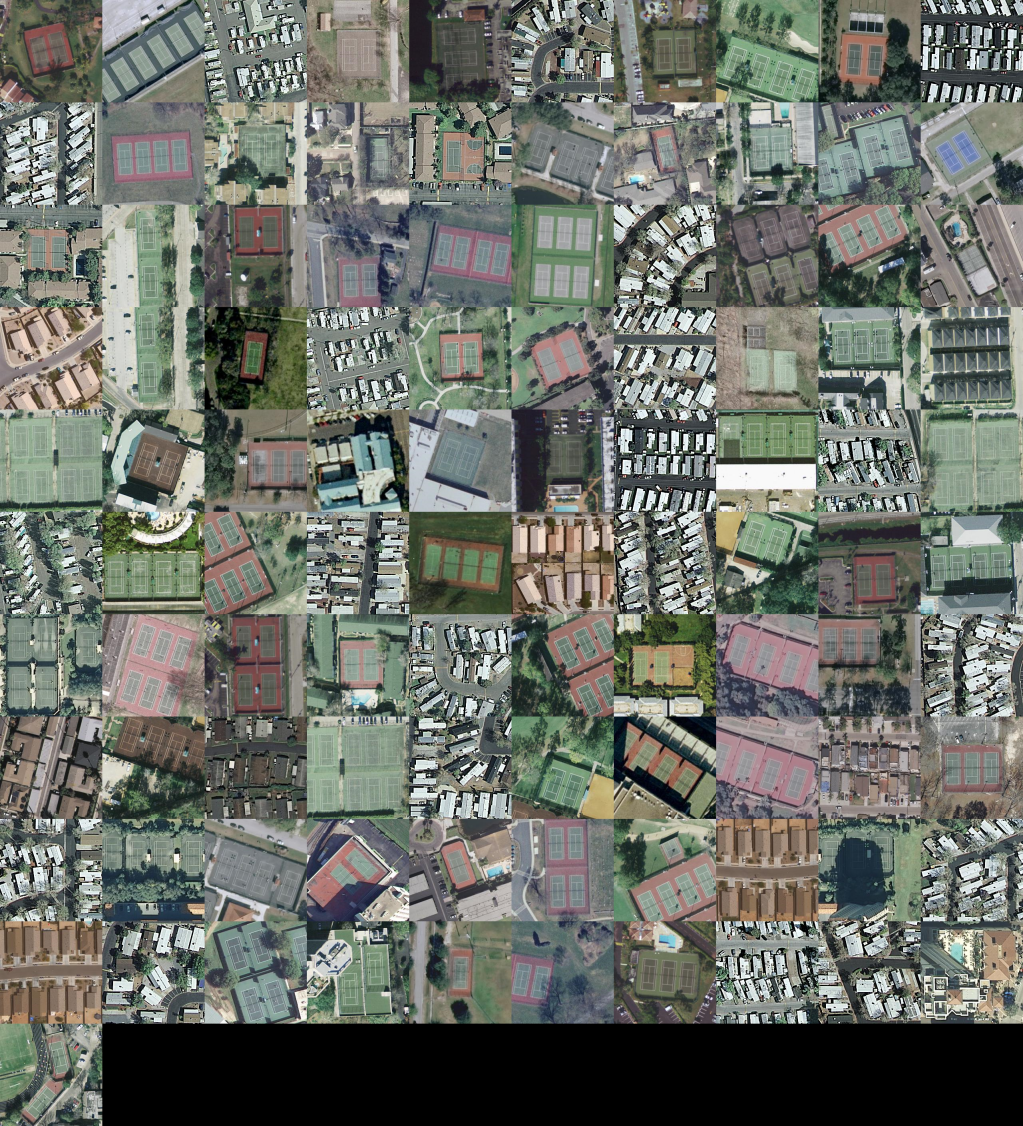}
 \caption{The remote--sensing images of tennis courts clustered by the clustering--based method \cite{li2021designing}. There are 101 images in total, and 71 images contain a tennis court.}
 \label{tenis_rp}
\end{figure}

When we apply the clustering--based representation learning method \cite{li2021designing} to UCMerced LandUse, the cluster containing most tennis court images is given in Figure \ref{tenis_rp}. We see that besides tennis court, there are some images of denser residential, building, and so on. There are 101 images in Figure \ref{tenis_rp}, and 71 of them contain a tennis court.

\begin{figure}[H] 
\centering

\subfigure {
 \label{fig:a}
\includegraphics[width=1.0\columnwidth]{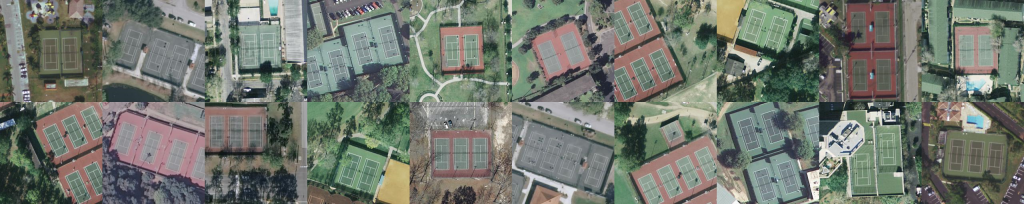}
}
(\textbf{a}) The images are ranked as the first group (the most reliable).

\subfigure {
\includegraphics[width=1.0\columnwidth]{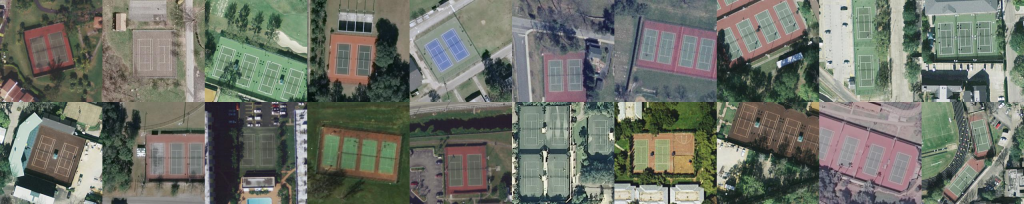}
}
(\textbf{b}) The images are ranked as the second group.

\subfigure {
\includegraphics[width=1.0\columnwidth]{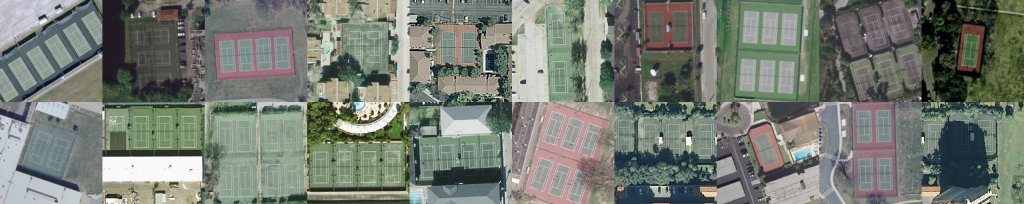}
}
(\textbf{c}) The images are ranked as the third group.

\subfigure {
\includegraphics[width=1.0\columnwidth]{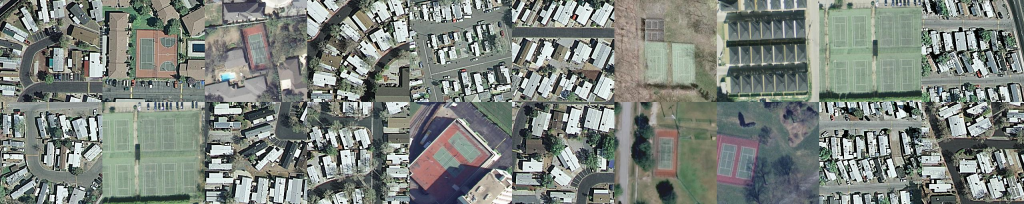}
}
(\textbf{d}) The images are ranked as the fourth group.

\subfigure {
\includegraphics[width=1.0\columnwidth]{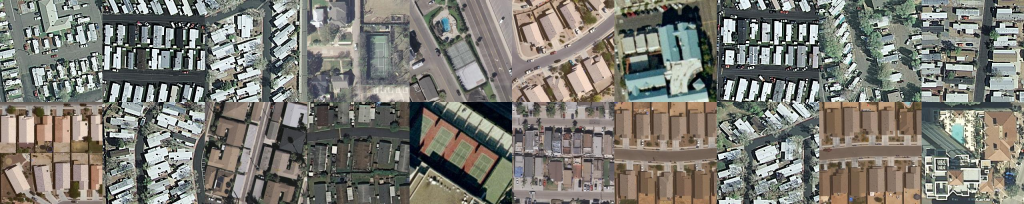}
}
(\textbf{e}) The images are ranked as the fifth group.

\caption{In the first group, that is the most reliable group of images containing a tennis court, all the images are correctly clustered. The images ranked in the second and third group are also all correctly clustered. In the fourth group, there are 1 image of dense residential, 11 images of mobile home parks, and 8 images of tennis courts. In the last group, there are only 3 images of tennis courts.
}
\label{tennis_ranking}
\end{figure}

Performing our ranking algorithm, we rank the images in Figure \ref{tenis_rp} into five groups, as shown in Figure \ref{tennis_ranking}. As can be seen from Figure \ref{tennis_ranking}, the images in the first, second, and third groups are all correctly clustered. There are still eight images of tennis courts in the fourth group and only three images of tennis courts in the last group. The ranking in Figure \ref{tennis_ranking} is expected: from the first to the last group, the reliability decreased. Other examples are given in \textbf{Appendix A, B}. 

\subsection{Ablation Study}
\label{ablation}
\begin{figure}[H]
\centering
 \includegraphics[width=9.6cm]{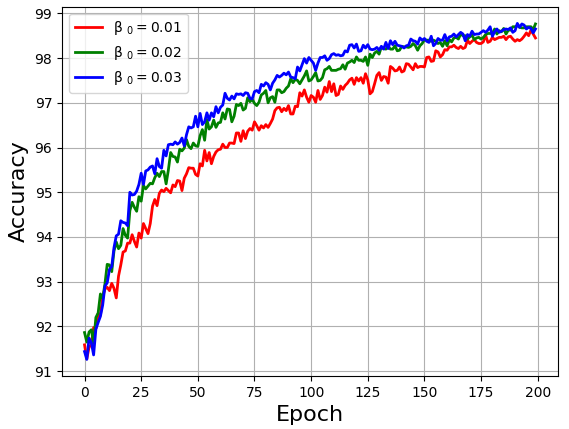}
 \caption{
Training behaviors of the ICSR technique for different values of the free parameter $\beta_0$ on SLT-10 clustering models from SPICE. It is seen that for different selections of the value $\beta_0$, training will converge to almost the same state after 200 epochs.
 }\label{acc_vs_beta}
\end{figure}

The ICSR technique has an important free parameter $\beta_0$ in (\ref{wi}), which controls the weights $w_i(t)$. 
In the training, we expect that $w_i(t)$ with $i> 1$ gradually increases. One reason for this is for seeking better weights, and another reason is that as training increases clustering accuracy, more samples are clustered correctly; thus, more pseudo labels become more reliable. In this paper, we rank samples in each cluster into five groups, and thus, there are five weights in (\ref{loss_w}). To investigate the effect of setting $\beta_0$, Figure \ref{acc_vs_beta} shows the training behaviors for three different values of $\beta_0$ on SLT-10 clustering models from SPICE (these accuracies were evaluated only on testing samples and, thus, higher than the accuracy in Table \ref{com}). It is seen that although the convergence behaviors of the models varied with different values, they all converged to the same state after 200 epochs, demonstrating that our method is not overly dependent on the selection of the free parameter and different values of the free parameter will lead to similar performances.
\unskip
\begin{figure}[H]
\centering
 \includegraphics[width=9.6cm]{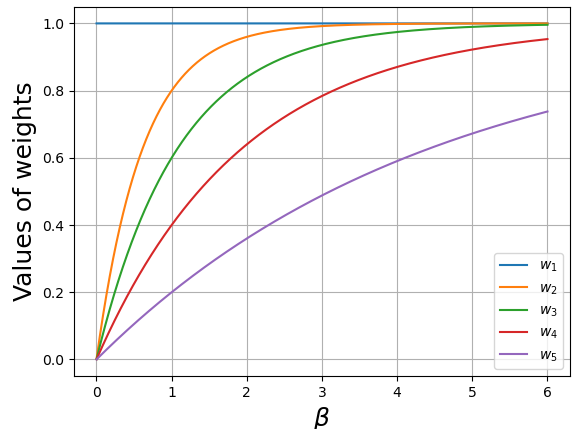}
 \caption{The values of $w_1,\;w_2,\;w_3,\;w_4,\;w_5$ given by $\beta$ increases from 0 to 6, where $\beta = (1+t)\beta_0$ (see Equation~(\ref{wi})).
 }\label{beta}
\end{figure}

In Figure \ref{beta}, the values of the five weights varied with $\beta$ ($\beta= (1+t)\beta_0$; see Equation~(\ref{wi})) are given. As expected, $w_1$ always keeps the same value, 1, unaffected by $\beta$, and $w_2>w_3>w_4>w_5$ is always correct for all $\beta$. When $\beta$ is very small, the values of $w_2,\;w_3,\;w_4,\;w_5$ are very close to zero, which approximately only keep the data of the first group. From Figure \ref{beta}, we see that $w_2$ grows much faster than $w_3,\;w_4,\;w_5$ as $\beta$ increases. In the phase that $w_2$ is very larger than $w_3,\;w_4,\;w_5$, there are approximately only two groups of data left. Similarly, when $w_3$ grows large enough, there are approximately only three groups of images left. Note that when $\beta_0=0.01$, $\beta$ increase from 0.01 to 1, we need 100 epochs, which means that there are many training epochs, which is roughly equivalent to that there are fewer groups of images. In other words, when the number of ranking groups is set as five, assigning weights as Equation~(\ref{wi}) roughly includes the cases with group numbers less than five.

From Figure \ref{acc_vs_beta}, we see that from Epoch 100 to Epoch 200, the three accuracies still increase, which correspond to $\beta$ from 1 to 2, 2 to 4, and 3 to 6, respectively. Note that the values of the five weights increased large enough after $\beta>1$, and thus, the increasing accuracies after epoch 100 imply that the samples or the pseudo labels that are not the most reliable also have made contributions to improve the performance of clustering. Recall that adding the samples or pseudo labels that are not the most reliable to training the model is the essential difference between our method and previous ones, such as SCAN~\cite{conf/eccv/GansbekeVGPG20}, RUC~\cite{park2021improving}, and SPICE~\cite{niu2021spice}.


\section{Conclusions}
Unsupervised image classification is challenging, but very important, as it can make use of abundant unlabeled data. This paper developed an effective image clustering technique that builds on the pretrained clustering models and improves their performances. \replaced[comment={replaced}]{To develop this approach, we solved the following problems: first of all, how to estimate the likelihoods of the samples belonging to their current clusters; secondly, how to improve the reliability of this estimation; thirdly, how to dynamically determine the contributions of the pseudo labels according to their confidence, which is varied with the training epochs. Based on the statistics, we introduced the quantity $M$ in Equation~(\ref{criterion}) to measure pseudo labels' confidence and employed majority voting to enhance the reliability of this measurement. A scheme is then proposed in Equation~(\ref{wi}) to weight the cross-entropy loss according to the confidence of the samples.} {The core of the new ICSR technique is a sample ranking and robust majority voting algorithm which ranks samples within clusters based on the likelihoods of the samples belong to their current clusters and use the samples ranks to weight their contributions to the cross-entropy loss function. We have presented extensive experimental results which demonstrated that the new ICSR method can achieve significant performance gain over State--of--the--Art models, even for those have already achieve very high performances.} \replaced[comment={replaced}] {We
applied the proposed method to various remote sensing datasets and compared the performance with SOTA methods. Through the results, we found that our method significantly outperforms SOTA methods in most cases.} {Applying our method to various remote sensing datasets shows that our method could be effectively used in the remote--sensing images} \added[comment={added}]{Besides clustering, the methods proposed in this paper could be applied to data denoising or data cleaning, automatic labeling, feature extracting, model pretraining, and so on.}
\vspace{6pt}

\appendix

\section{The Performance of Ranking Samples of Freeway in UCMerced LandUse}

\begin{figure}[H]
\centering
 \includegraphics[width=12.8cm]{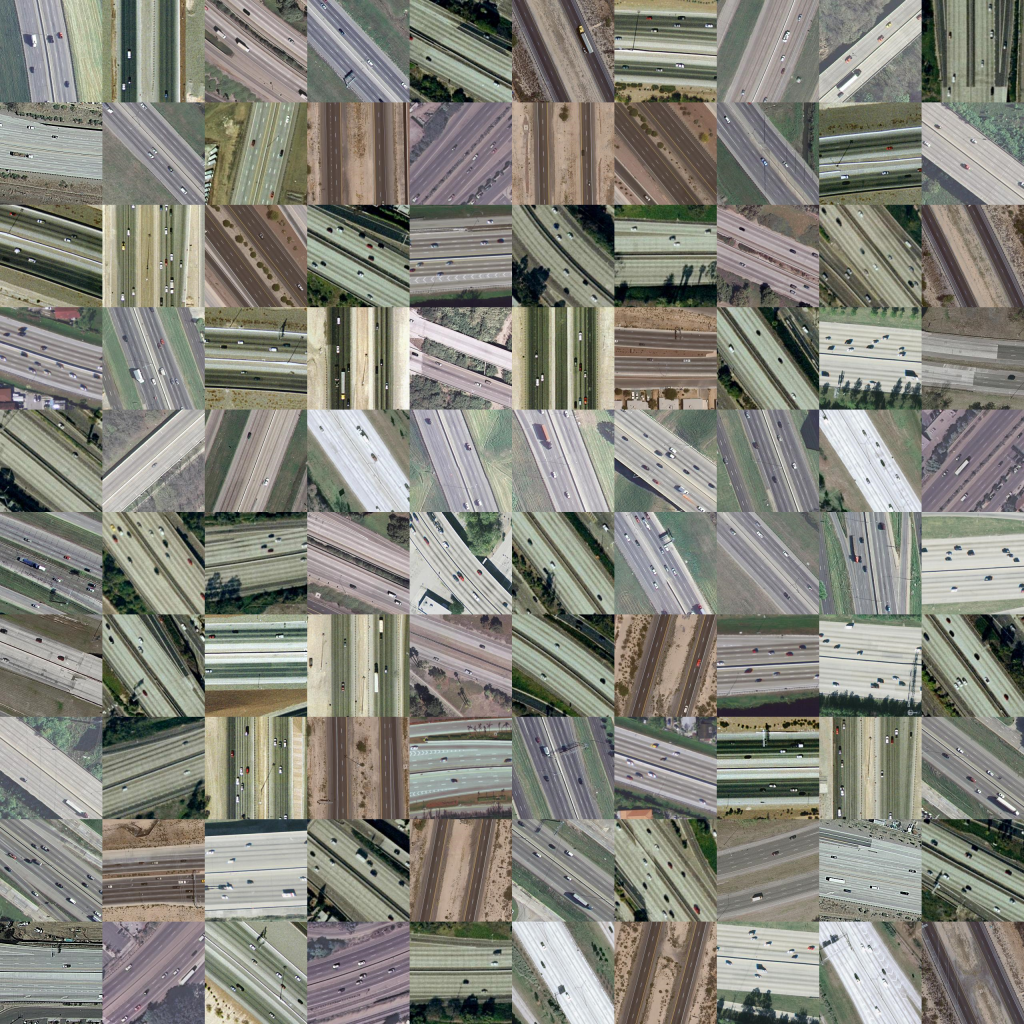}
 \caption{Ground truth for remote--sensing images of freeway. There are 100 images of freeway in total.}
 \label{greeway_groundtruth}
\end{figure}
\unskip

\begin{figure}[H]
\centering
 \includegraphics[width=12.8cm]{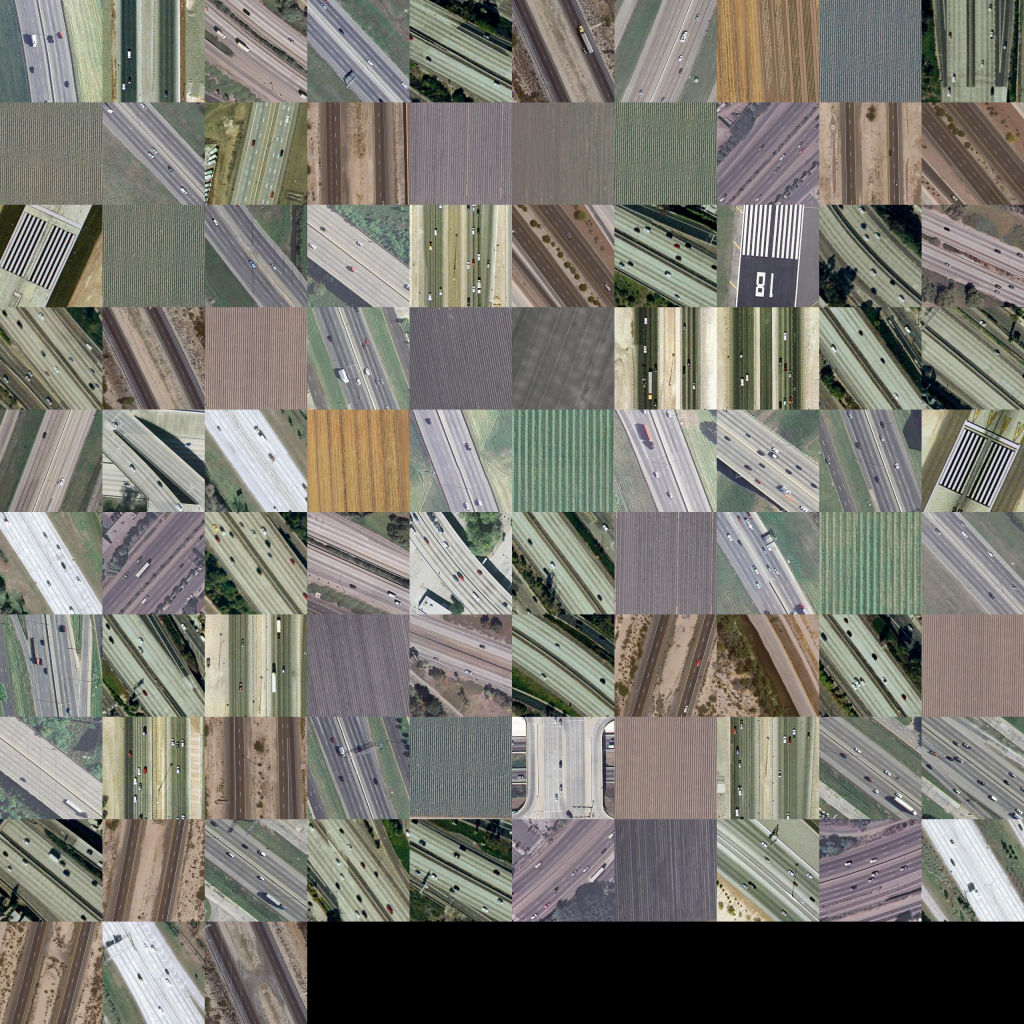}
 \caption{The remote--sensing images of freeway clustered by clustering--based method \cite{li2021designing}. There are 93 images in total, and 68 images contain freeway.}
 \label{feeway_rp}
\end{figure}

\begin{figure}[H] 
\centering

\subfigure {
 \label{fig:a}
\includegraphics[width=1.0\columnwidth]{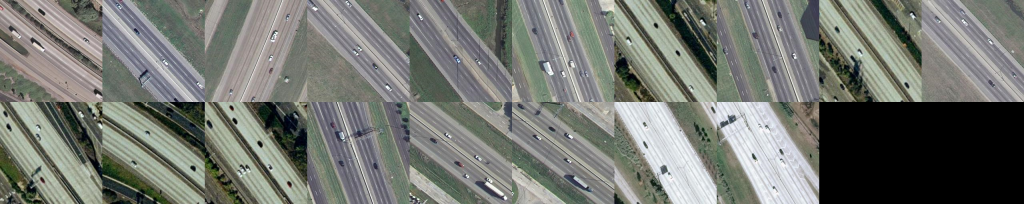}
}
(\textbf{a}) The images are ranked as the first group (the most reliable).

\subfigure {
\includegraphics[width=1.0\columnwidth]{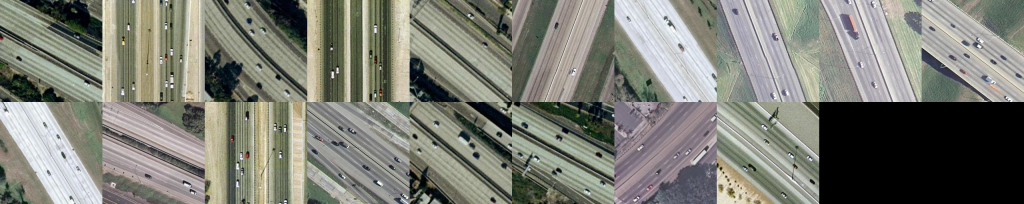}
}
(\textbf{b}) The images are ranked as the second group.

\subfigure {
\includegraphics[width=1.0\columnwidth]{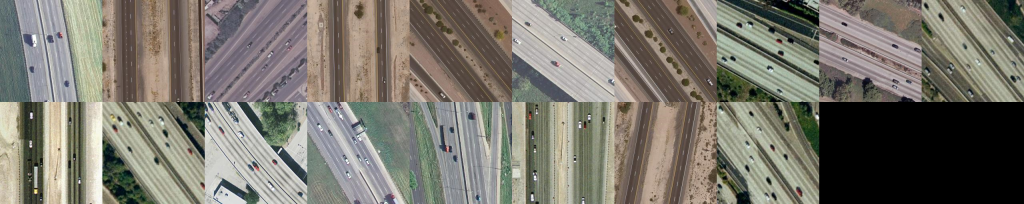}
}
(\textbf{c}) The images are ranked as the third group.

\subfigure {
\includegraphics[width=1.0\columnwidth]{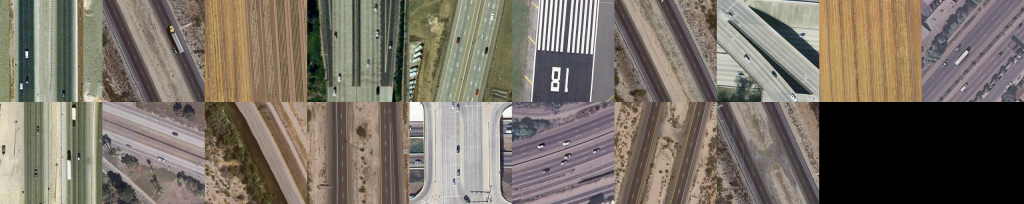}
}
(\textbf{d}) The images are ranked as the fourth group.

\subfigure {
\includegraphics[width=1.0\columnwidth]{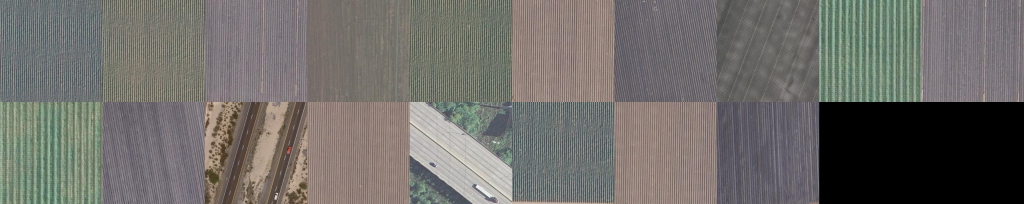}
}
(\textbf{e}) The images are ranked as the fifth group.

\caption{In the first, second, and third groups, all the images are correctly clustered. In the fourth group, there is 1 image of river and runway, 2 images of agricultural and overpass, and 12 images of freeway. In the last group, there are only 2 images of freeway.
}
\label{freeway_ranking}
\end{figure}

\section{The Performance of Ranking Samples of Beach in UCMerced LandUse}

\begin{figure}[H]
\centering
 \includegraphics[width=12.8cm]{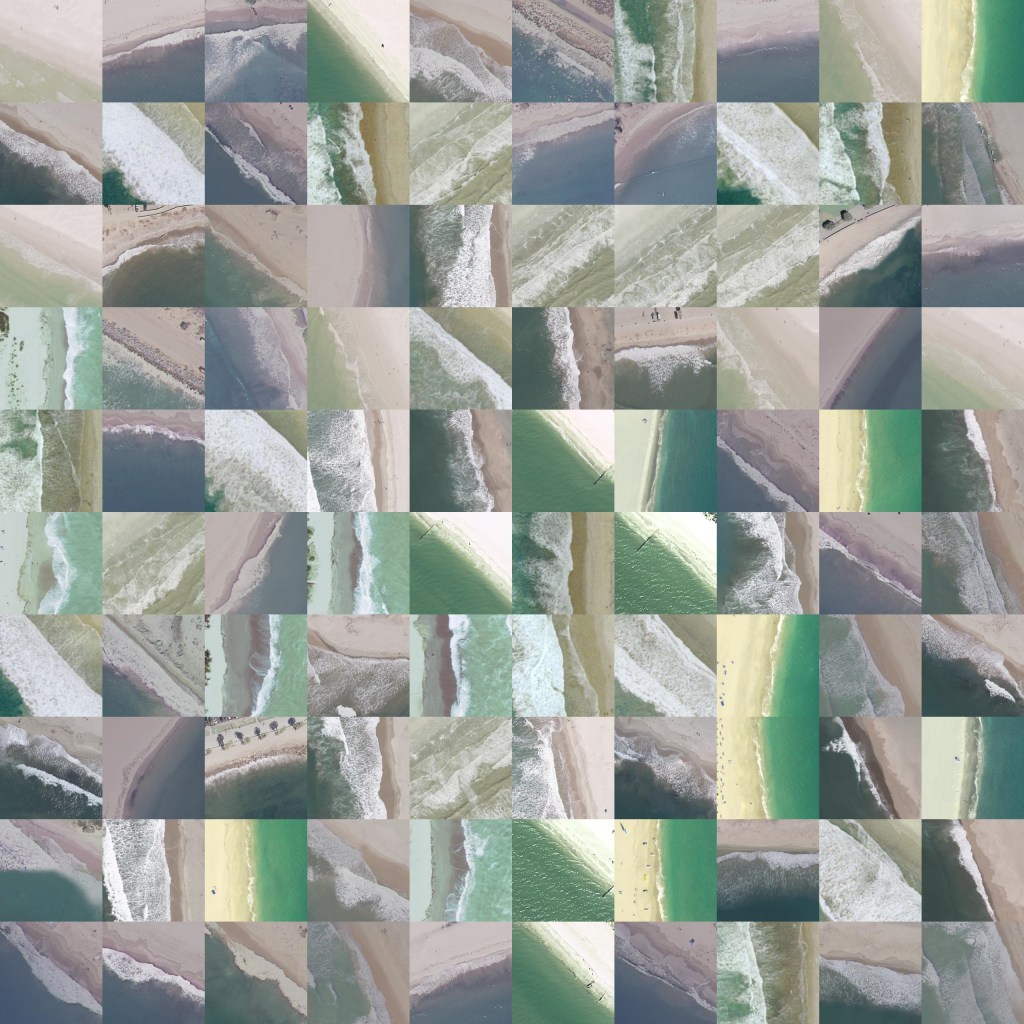}\\
 \caption{Ground truth for remote--sensing images of beach. There are 100 images of beach in total.}
 \label{beach_groundtruth}
\end{figure}
\unskip

\begin{figure}[H]
\centering
 \includegraphics[width=12.8cm]{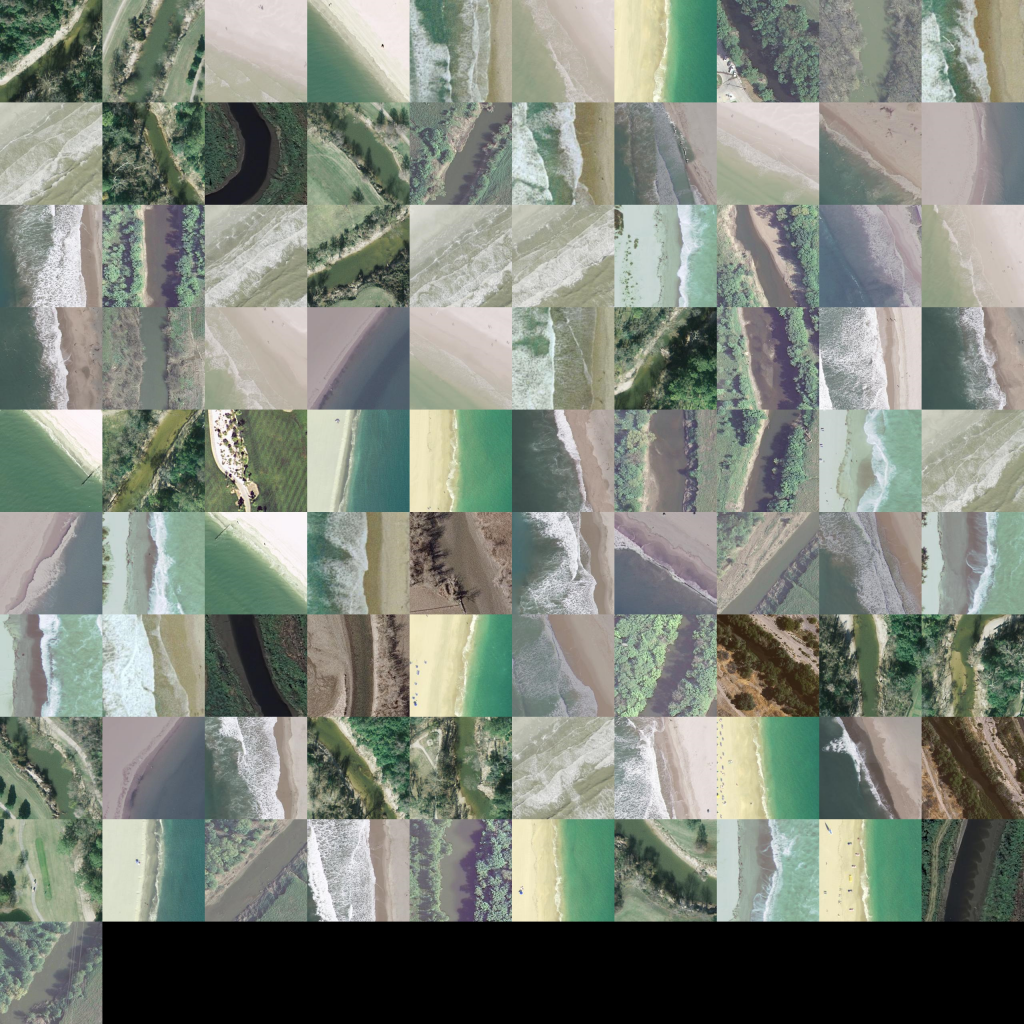}\\
 \caption{The remote--sensing images of beach clustered by the clustering--based method \cite{li2021designing}. There are 101 images in total, and 60 images contain beach.}
 \label{beach_rp}
\end{figure}

\begin{figure}[H] 
\centering

\subfigure {
 \label{fig:a}
\includegraphics[width=1.0\columnwidth]{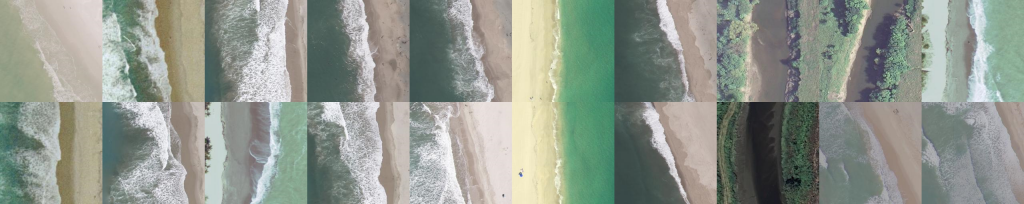}
}
(\textbf{a}) The images are ranked as the first group (the most reliable).

\subfigure {
\includegraphics[width=1.0\columnwidth]{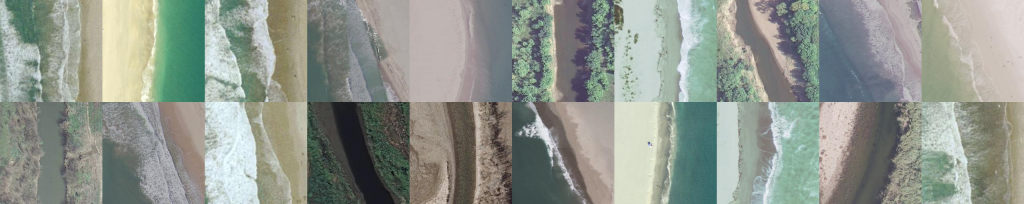}
}
(\textbf{b}) The images are ranked as the second group.

\subfigure {
\includegraphics[width=1.0\columnwidth]{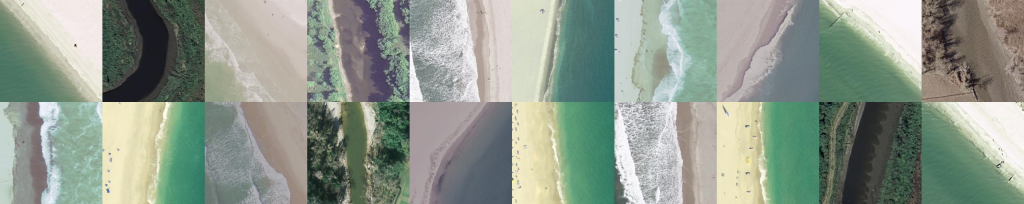}
}
(\textbf{c}) The images are ranked as the third group.

\subfigure {
\includegraphics[width=1.0\columnwidth]{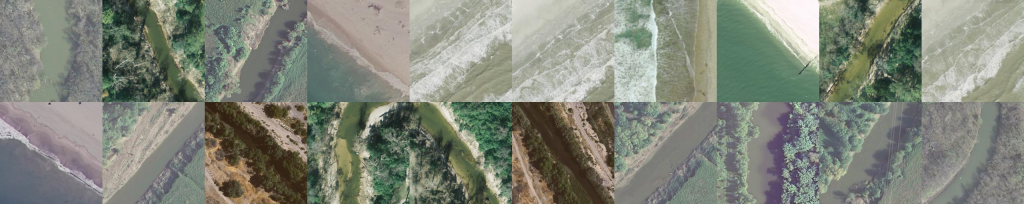}
}
(\textbf{d}) The images are ranked as the fourth group.

\subfigure {
\includegraphics[width=1.0\columnwidth]{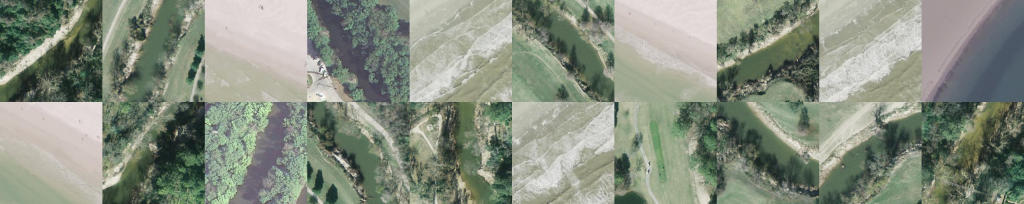}
}
(\textbf{e}) The images are ranked as the fifth group.

\caption{In the first group, there are 3 images of river and 17 images of beach. In the second group, there are 6 images of river and 14 images of beach. In the third group, there are 5 images of river and 15 images of beach. In the fourth group, there are 13 images of river and 7 images of beach. In the last group, there are 1 image of golf course, 12 images of river, and 7 images of beach. 
}
\label{beach_ranking}
\end{figure}


\end{document}